\documentclass{article}

\usepackage{microtype}
\usepackage{graphicx}
\usepackage{subfigure}
\usepackage{booktabs} 

\usepackage{mathtools}
\usepackage{stmaryrd}
\usepackage[toc,page]{appendix}
\usepackage{times}
\usepackage{epsfig}
\usepackage{graphicx}
\usepackage{amsmath}
\usepackage{caption}
\usepackage{amssymb}
\usepackage{amsthm}
\usepackage{ifthen}
\usepackage{caption,booktabs,array}
\usepackage{multirow}
\usepackage{dsfont}
\usepackage{tikz}
\usepackage{xcolor}
\usepackage{comment}

\usepackage{times}
\usepackage{color}
\definecolor{webred}{rgb}{0.5,0,0}

\newcommand\V{\vert}
\renewcommand{\a}{\alpha}

\usepackage[utf8]{inputenc}

\def\aa{{\boldsymbol a}}
\def\bb{{\boldsymbol b}}
\def\uu{{\boldsymbol u}}
\def\xx{{\boldsymbol x}}
\def\yy{{\boldsymbol y}}
\def\zz{{\boldsymbol z}}

\def\XX{{\boldsymbol X}}
\def\YY{{\boldsymbol Y}}

\def\R{{\mathbb{R}}}
\def\PP{{\mathcal P^m}}

\def\expect{\mathop{\mathbb{E}}}
\newtheorem{definition}{Definition}
\newtheorem{theorem}{Theorem}

\newtheorem{lemma}{Lemma}
\newtheorem{proposition}{Proposition}

\usepackage[pagebackref=true,breaklinks=true,letterpaper=true,bookmarks=false]{hyperref}

\newcommand{\defas}{\;\mathrel{\!\!{:}{=}\,}}

\newtheorem{innercustomgeneric}{\customgenericname}
\providecommand{\customgenericname}{}
\newcommand{\newcustomtheorem}[2]{%
  \newenvironment{#1}[1]
  {%
   \renewcommand\customgenericname{#2}%
   \renewcommand\theinnercustomgeneric{##1}%
   \innercustomgeneric
  }
  {\endinnercustomgeneric}
}
\newcustomtheorem{customdef}{Definition}
\newcustomtheorem{customtheorem}{Theorem}
\newcustomtheorem{customcorollary}{Corollary}
\newcustomtheorem{customexample}{Example}
\newcustomtheorem{customlemma}{Lemma}
\newcustomtheorem{customprop}{Proposition}

\usepackage{hyperref}

\usepackage[accepted]{icml2021}

\icmltitlerunning{Unbalanced minibatch Optimal Transport; applications to Domain Adaptation}

\begin{document}

\twocolumn[
\icmltitle{{Unbalanced minibatch Optimal Transport; applications to Domain Adaptation}}




\begin{icmlauthorlist}
\icmlauthor{Kilian Fatras}{to}
\icmlauthor{Thibault S\'ejourn\'e}{goo}
\icmlauthor{Nicolas Courty}{to}
\icmlauthor{R\'emi Flamary}{ed}
\end{icmlauthorlist}

\icmlaffiliation{to}{Univ. Bretagne-Sud, CNRS, \textsc{Inria}, IRISA, France}
\icmlaffiliation{goo}{ENS, PSL University}
\icmlaffiliation{ed}{\'Ecole Polytechnique, CMAP, France}

\icmlcorrespondingauthor{Kilian Fatras}{kilian.fatras@irisa.fr}

\icmlkeywords{Machine Learning, ICML}

\vskip 0.3in
]



\printAffiliationsAndNotice{} 


\begin{abstract}

{Optimal transport distances have found many applications in machine learning
for their capacity to compare non-parametric probability distributions. Yet
their algorithmic complexity generally prevents their direct use on large scale
datasets. Among the possible strategies to alleviate this issue, practitioners
can rely on computing estimates of these distances over subsets of data, {\em
i.e.} minibatches. While computationally appealing, we highlight in this paper
some limits of this strategy, arguing it can lead to undesirable smoothing
effects. As an alternative, we suggest that the same minibatch strategy coupled
with unbalanced optimal transport can yield more robust behavior. We discuss
the associated theoretical properties, such as unbiased estimators, existence of
gradients and concentration bounds. Our experimental study shows that in
challenging problems associated to domain adaptation, the use of unbalanced
optimal transport leads to significantly better results, competing with or
surpassing recent baselines.
}
\end{abstract}

\section{Introduction}

{Computing} distances between distributions is a fundamental problem in machine
learning. As an example, considering the space of distributions $\mathcal{M}_{+}(\mathcal{X})$ over a
space $\mathcal{X}$, and given an empirical distribution $\alpha \in
\mathcal{M}_{+}(\mathcal{X})$, {many machine learning problems amount to
estimate} a  distribution $\beta_{\lambda}$ parametrized by a vector $\lambda$
which approximates the distribution $\alpha$. In order to {compute} the
dissimilarities between distributions, {it is common to rely on a contrast
function or divergence} $L : \mathcal{M}_{+}(\mathcal{X}) \times
\mathcal{M}_{+}(\mathcal{X}) \to \mathbb{R}_+$. {In this setting}, the goal
is to find the optimal {$\lambda^*$} which minimizes the distance $L$ between
the distributions $\beta_{\lambda}$ and $\alpha$, {{\em i.e.} $ \lambda^* =
\operatorname{argmin}_\lambda L(\alpha, \beta_{\lambda})$}. As the available
distributions are {mostly} empirical {and come from data}, {the
function $L$ needs} good statistical estimation properties and optimization
guarantees when using modern optimization techniques. Optimal transport (OT) losses
have emerged recently as a competitive loss candidate for generative models
\cite{arjovsky_2017,genevay_2018}. It also proved to be competitive in the
context of Domain Adaptation
\cite{DACourty, Courty2017,shen2018wasserstein} or for missing data imputation \cite{muzellec2020missing}. The corresponding estimator is usually found in the
literature under the name of {\em Minimum Kantorovich
Estimator}~\cite{Bassetti06,COT_Peyre}. However the computation of OT losses is
a challenging problem, its computational cost being of
order $\mathcal{O}(n^3log(n))$, where $n$ is the number of samples.
Variants and approximations of optimal transport have been proposed to reduce its complexity. One of the most popular consists in adding an entropic regularization~\cite{CuturiSinkhorn}, leading to the Sinkhorn algorithm with complexity $\tilde{\mathcal{O}}(n^2)$ in both space and time.
However, when $n$ is large, computing OT remains rather expensive and might not fit on GPUs. The KeOps package~\cite{feydy19a} allows to overcome this difficulty and avoid overflows by storing operations as formulas and stream computation on the fly. It is still difficult to use it in deep learning applications which involves high dimensional data and repeated computations of gradients. Another approach is to focus on the Wasserstein-1 distance which has a nice reformulation but needs to approximate 1-Lipschitz functions, which meets some difficulties in practice~\cite{arjovsky_2017, Gulrajani2017}. 

\begin{figure}[!t]
    \centering
    \includegraphics[scale=0.37]{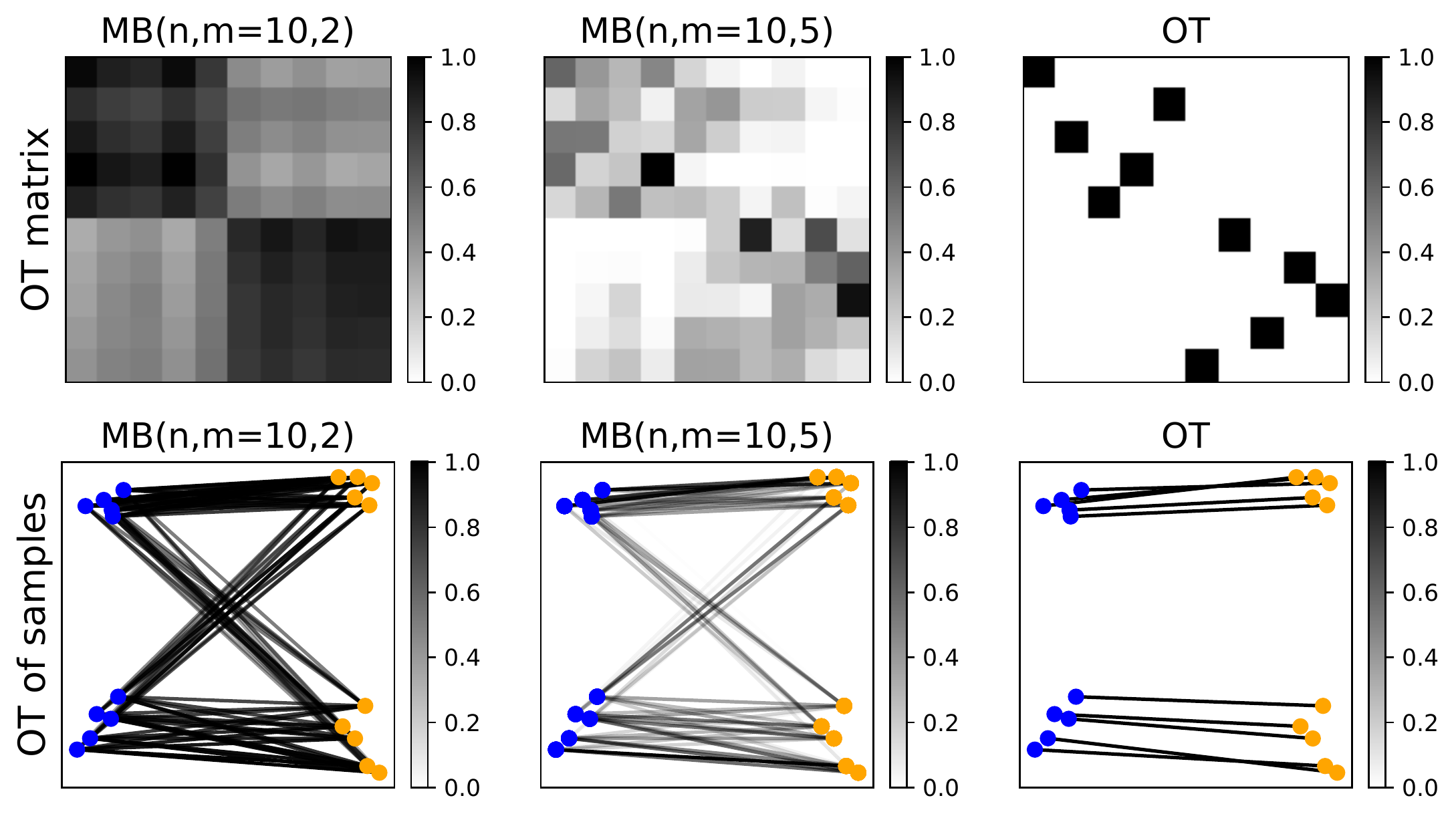}
    \vspace{-0.7cm}
    \caption{OT matrices, normalized by maximum value, between 2D distributions with $n=10$ samples. The first row shows the MBOT plans for different minibatch size $m$. The second row is the corresponding 2D visualization of the transport plan support.\label{fig:2D_gauss}}
    \label{fig:mbot_non_opt_connection}
\end{figure}

{\bfseries{Minibatch Optimal Transport.} } 
A straightforward and scalable approach consists in computing OT solutions over
subsets (minibatches) of the original data ($\alpha$ and $\beta$) and averaging
the results as a proxy for the original problem. 
Such idea stems from the need to scale OT in practice and was applied in several situations~\cite{kolouri2016sliced,genevay_2018,deepjdot, liutkus19a}.
It was proven for generative models that minimizers of the minibatch loss converge to the true minimizer when the minibatch size increases~\cite{bernton2017}. 
There exists deviation bounds between the true OT loss and a single minibatch estimate~\cite{mbot_Sommerfeld}. Finally, concentration bounds and optimization properties for averaged minibatch OT were exhibited in~\cite{fatras20a, fatras2021minibatch}. 
However, the gain in computation time {is achieved at the expense of the quality of the final transport plan, which turns out to be notably less sparse}
, leading to undesired pairings between samples that would not be coupled with exact OT. 
Figure~\ref{fig:mbot_non_opt_connection} illustrates this effect on a 2D toy example which shows that samples from the same cluster in the source probability can be coupled to two different clusters in the target.  
We propose to handle this problem by leveraging the theory of unbalanced OT and computing a more robust transport plan at the minibatch level.

{\bfseries{Contributions and outline of the paper.}}
We study in this work an alternative formulation of the minibatch OT where the unbalanced OT program, a variant with relaxed marginal constraints~\cite{Liero_2017}, is used at the minibatch level. 
Our rationale is that a geometrically robust version of OT computed between minibatches decreases the influence of undesired couplings between samples.
The benefits of unbalanced MBOT are twofold: 
{\em i)} it yields a loss function more robust to minibatch sampling effects
{\em ii)} our formulation approximates unbalanced OT but scales computationally w.r.t. the minibatch size, which allows its practical use for large datasets and deep learning applications. 
The contributions of the paper are the following. First we review
the existing UOT formulations and introduce the one we consider in Section~\ref{sec:rw}. 
We discuss the limits of minibatch OT in Section~\ref{sec:robust}.
We present the minibatch
framework, study its statistical and
optimization properties in Section~\ref{sec:mbuot}. 
Finally, we design a new domain adaptation (DA) method whose performances are
evaluated on several problems, where we show evidences that our strategy surpasses
substantially other classical OT formulations, and is on par or better than
recent state-of-the-art competitors. {Our empirical results suggest that UOT might be more suitable than OT when dealing with real world data.}

{\bfseries{Notations.}}
In this paper, we use the following notations. Let $\boldsymbol{X}=(\xx_1, \cdots, \xx_n) $ (resp. $\boldsymbol{Y}=(\yy_1, \cdots, \yy_n) $) be $n$ \emph{iid} random vectors in $\R^d$ drawn from a distribution $ \alpha $ (resp. $ \beta $) on the source (resp. target) domain. We associate to $ \{\xx_1, \cdots, \xx_n \} $ and  $ \{ \yy_1, \cdots, \yy_n \} $ uniform vectors denoted $(\uu_n)_i = (\frac{m_\alpha}{n})_{i}$, where $m_\alpha = \int d\alpha$ is the mass of $\alpha$. The quantities $\XX$ and $\uu_n$ allows one to recover an empirical distributions as $\alpha_n = \frac1n \sum_i  \delta_{\xx_i}$. We denote $\alpha^{\otimes m }$ for a sample of $m$ random variables following the distribution $\alpha$. The ground cost can be formalised as the following map:
\begin{equation}\label{def:ground_cost}
C^n:(\XX,\YY) \mapsto \big(d(\xx_i, \yy_j)\big)_{\small{1\leq i,j\leq n}} \in \mathcal{M}_n(\R).
\end{equation}
We further suppose that $\alpha$ and $\beta$ have compact support which means that the ground cost is bounded by a strictly positive constant $M$. This assumption holds for most machine learning applications where distributions are given by empirical samples.

\long\def\comment#1{\paragraph{Robust Optimal Transport.} Robustness in optimal transport  mostly
tries to mitigate the noise in high dimensional data, and can be understood in
many ways. The first line of work considers this problem under the perspective
of learning a ground cost which maximises the OT distance, translating into a
max-min problem~\cite{genevay_2018,paty2019subspace,dhouib2020swiss}. This
includes for instance subspace seeking \cite{paty2019subspace} or optimization
over a compact set of cost matrices~\cite{dhouib2020swiss}. Another line of work
relaxes the marginal constraints by introducing a min-max problem which uses the dual formulation of OT~\cite{balaji2020robust}, or by adding a regularization on input measures~\cite{mukherjee2020outlierrobust}.

}

\section{Related work and background}\label{sec:rw}

{We review in this section previous Unbalanced OT formulations, detail the one we consider in our approach and discuss the use of OT in robust machine learning. }

{\bfseries{Unbalanced Optimal Transport.}}
{Unbalanced OT is a generalization of 'classical' OT that relaxes the conservation of mass constraints by allowing the system to either transport or create and destroy mass.
Our loss builds upon~\cite{Liero_2017} which replaces the 'hard' marginal constraints of OT by 'soft' penalties using Csisz\`ar divergences. 
There exists other extensions of the static formulations of OT. A famous one is partial OT which consists in transporting a fixed budget of mass~\cite{figalli_partial} or to move mass in and out of the system at a fixed cost~\cite{figalli2010new}. Another line of work proposes to optimize over various sets of Lipschitz functions~\cite{Hanin1992,piccoli2014,Schmitzer2017}. One can also replace Csiszàr divergences by integral probability metrics~\cite{nath2020unbalanced}.

{Consider a convex, positive, lower-semicontinuous function $\phi$ such that $\phi(1)=0$. Define $\phi^\prime_\infty = \lim_{x\rightarrow +\infty} \phi(\xx) / \xx$ that we suppose strictly positive. Csiszàr divergences $D_\phi$ are measures of discrepancy that compare pointwise ratios of mass using a penalty $\phi$ and are defined as $D_\phi(\xx, \yy) = \sum_{\yy_i\neq 0} \yy_i \phi\Big( \frac{\xx_i}{\yy_i} \Big) + \phi^\prime_\infty \sum_{\yy_i = 0} \xx_i$. Total Variation and Kullback-Leibler divergences ($ \texttt{KL}(\xx|\yy) = \sum_{i} \xx_{i}\log(\frac{\xx_{i}}{\yy_i}) - \xx_{i} + \yy_i$) are particular instances of such divergence.
Consider two positive distributions $\alpha, \beta \in \mathcal{M}_{+}(\mathcal{X})$. The UOT program between distributions and cost $c$ is
defined as}
\begin{align}
\label{eq-uot-def}
  \operatorname{OT}_\phi^{\tau, \varepsilon}(\alpha, \beta, c) &= \underset{\pi \in \mathcal{M}_+(\mathcal{X}^2)}{\text{min}} \int cd\pi + \varepsilon \texttt{KL}(\pi|\alpha \otimes \beta)\nonumber \\
  &\quad \qquad {+ \tau (D_\phi(\pi_1\|\alpha)  + D_\phi(\pi_2\|\beta))},
\end{align}
where $\pi$ is the transport plan, $\pi_1$ and $\pi_2$ the plan's marginals, $\tau$ is the marginal penalization and $\varepsilon \geq 0$ is the regularization coefficient. Note that the marginals of $\pi$ are no longer equal to $(\alpha,\beta)$ in general.}
The considered formulation is computable via a generalized Sinkhorn algorithm~\cite{ChizatPSV18, sejourne2019sinkhorn} which is proved to converge. Its complexity for $D_\phi=\texttt{KL}$ is $\tilde{O}(n^2/\epsilon)$~\cite{pham20a}.
{Balanced OT is
recovered for inputs $(\alpha,\beta)$ with equal mass, when $\tau \rightarrow
\infty$ (hence we note it $\operatorname{OT}_\phi^{\infty, 0}$).} When distributions are discrete, UOT can be expressed as $\operatorname{OT}_\phi^{\tau, \varepsilon}(\aa, \bb, C)$ where $\aa$ and $\bb$ are two positive vectors, $\aa,\bb \in \mathbb{R}_+^n$ and $C$ is the ground cost.

A shortcoming of adding entropy is the loss of metric properties since $\operatorname{OT}_\phi^{\tau,
\varepsilon}(\beta, \beta, c) \ne 0$. {It motivated \cite{sejourne2019sinkhorn} to introduce an unbalanced generalization of the Sinkhorn divergence~\cite{genevay_2018}}:
\begin{align}
S_\phi^{\tau, \varepsilon}(\alpha, \beta, c) &= \operatorname{OT}_\phi^{\tau, \varepsilon}(\alpha, \beta, c) + \frac{\varepsilon}{2}(m_\alpha - m_\beta)^2 \\
& \quad -\frac12 \operatorname{OT}_\phi^{\tau, \varepsilon}(\alpha, \alpha, c) - \frac12\operatorname{OT}_\phi^{\tau, \varepsilon}(\beta, \beta, c) \nonumber,
\end{align}
Computing the unbalanced sinkhorn divergence above is of the same order of complexity as the
UOT loss. When $e^{-C/\epsilon}$ is a positive definite kernel, $S_\phi^{\tau,\varepsilon}$ is a convex, symmetric, positive definite
loss function which metrizes the convergence in law~\cite{sejourne2019sinkhorn}. Thus it allows to mitigate between accelerated computations and conservation of key theoretical guarantees. 
Regarding empirical estimation, OT suffers from the curse of dimension which means that it is hard to estimate when data lie in high dimension $d$. Its sample complexity, \emph{i.e.,} its convergence in population, is proven to be in $O\left(\frac{1}{\sqrt{n}}\left(1+\frac{1}{\varepsilon^{\lfloor d / 2\rfloor}}\right)\right)$ both for OT and UOT~\cite{genevay19, sejourne2019sinkhorn}.


{\bfseries{Optimal Transport and robustness in machine learning.}}
UOT is known to be more robust to outliers than OT as it does not need to meet the marginals. Several other formulations make optimal transport robust for practical and statistical reasons.
Partial OT can be adapted for partial matchings problem with applications for positive-unlabeled learning~\cite{chapel2020partial}. 
A line of work proposes 'distributionnally robust' models, where models are trained in a Wasserstein ball around the empirical distribution in the space of probabilities~\cite{Esfahani_2018, Kuhn_2019}.
 In a similar approach, several variants relax the OT marginal constraints with a ball constraint, and consider several penalties such as integral probability metrics~\cite{nath2020unbalanced}, total variation or Csiszàr divergences for outlier detection~\cite{mukherjee2020outlierrobust, balaji2020robust}.
Such relaxations allow to derive statistical guarantees w.r.t. noise and outliers.
%
%
%
Another idea to ensure robustness consists in learning the cost adversarially, and is formulated as a
max-min problem where the cost is modeled by an Euclidean embedding~\cite{genevay_2018}, a compact space of matrices~\cite{dhouib2020swiss} or a projection on a lower dimensional subspace~\cite{paty2019subspace}. 

In the next section we discuss OT sensitivities in more details and highlight their exacerbation by minibatch strategy.  


%


\long\def\comment#1{\paragraph{Entropic regularization}
To fasten the computation of optimal transport and reduces its sample complexity, one can add to the optimal transport problem an entropic regularization on the transport plan\cite{CuturiSinkhorn}. Regularized entropic OT can be solved with the Sinkhorn algorithm which has a near-quadratic complexity in the size of samples\cite{altschuler2017near}. It can be defined for both balanced and unbalanced optimal transport and we give the unbalanced formulation. We define it as:
\begin{align}
  \operatorname{OT}^{\tau, \varepsilon}(\aa, \bb) &= \underset{\Pi \in \mathbb{R}_+^{n \times n}}{\text{min}} \langle C, \Pi \rangle + \varepsilon \texttt{KL}(\Pi|\aa \otimes \bb) \nonumber\\
  &\qquad + \tau (\texttt{KL}(\Pi\mathbf{1}_n\|\aa) + \texttt{KL}(\Pi^T\mathbf{1}_n\|\bb))
\end{align}
where $\varepsilon$ is the regularization coefficient. Adding an entropic regularization to the transport problem leads to a sub-optimal transport plan $\Pi$ of the original problem. Furthermore it breaks the metric property since $\operatorname{OT}_{c}^{\varepsilon}(\bb, \bb) \ne 0$. This motivated \cite{genevay_2018} to introduce an unbiased loss which uses the entropic regularization and called it the Sinkhorn divergence. It was recently extended for unbalanced OT in \cite{sejourne2019sinkhorn} and is defined as:
\begin{align}
S_\phi^{\tau, \varepsilon}(\aa, \bb) &= \operatorname{OT}_{c}^{\varepsilon}(\aa, \bb) -  \frac12(\operatorname{OT}_{c}^{\varepsilon}(\bb, \bb) + \operatorname{OT}_{c}^{\varepsilon}(\aa, \aa))\nonumber \\
& \quad +\frac{\varepsilon}{2}(m_\aa - m_\bb)^2,
\end{align}
with $m_\aa = \sum_{i=1}^n a_i$. Its computation cost is of the same order of complexity as the entropic loss. The balanced case has been proven to interpolate between OT and maximum mean discrepancy (MMD) \cite{sejourne2019sinkhorn, feydy19a} with respect to the regularization coefficient. MMD are integral probability metrics over a reproducing kernel Hilbert space \cite{MMD_Gretton}. When $\varepsilon$ tends to 0, we get the OT solution back and when $\varepsilon$ tends to $\infty$, we get a solution closer to the MMD solution. Second, as proved by \cite{feydy19a, sejourne2019sinkhorn}, if the cost $c$ is Lipschitz, then $S_{c}^{\varepsilon}$ is a convex, symmetric, positive definite loss function. Hence the use of the Sinkhorn divergence instead of the regularized OT should be advocated to avoid undesirable  effects of entropic regularization. The sample complexity of the Sinkhorn divergence, that is the convergence rate of a metric between a probability distribution and its empirical counterpart as a function of the number of samples, was proven in  \cite{genevay19, sejourne2019sinkhorn} to be:
$
O\left(\frac{e^{\frac{\kappa}{\varepsilon}}}{\sqrt{n}}\left(1+\frac{1}{\varepsilon^{\lfloor d / 2\rfloor}}\right)\right)
$ where $d$ is the dimension of $\mathcal{X}$. We see an interpolation between MMD and OT sample complexity depending on $\varepsilon$.
}


\section{Minibatch OT and robustness to sampling}\label{sec:robust}
 \begin{figure}[t]
     \centering
     \includegraphics[scale=0.275]{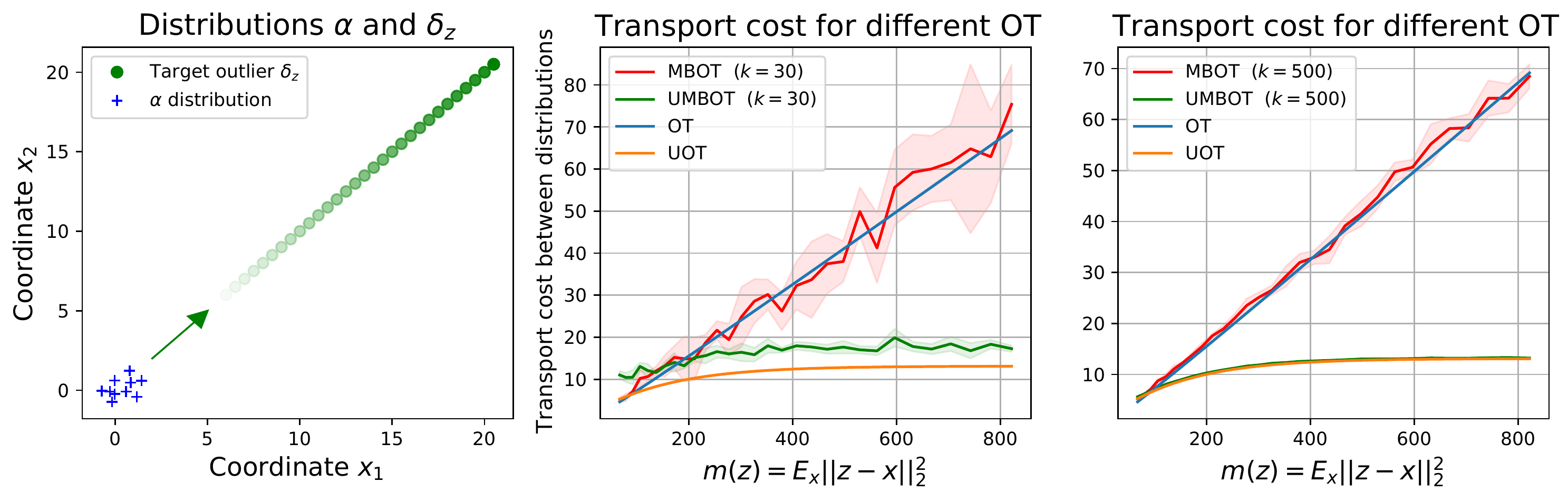}
     \vspace{-0.7cm}
     \caption{ Several OT costs between 2D distributions with $n=10$ samples and $m=5$. Target distribution is equal to the source distribution tainted with a moving outlier (green dot). The shaded area represent the variance of subsample MBOT on 5 run (see section \ref{sec:mbuot}).\label{fig:deviation}}
 \end{figure}

{
In this section, we discuss the limitations of combining balanced OT with the minibatch framework.
OT is sensitive to the distributions geometry. When those distributions are tainted by outliers, OT is forced to transport them due to the marginal constraints, inducing an undesirable extra transportation cost.  Minibatch OT averages several OT terms related to subsamples of the original distributions, thus sharing this sensitivity. The problem is even worse as two minibatches do not necessarily share samples that would lie in the support of the full OT plan, hence forced to match samples that could be, at the level of a minibatch, considered as outliers. Take as an example two distributions with clustered samples. While in the full OT plan clusters can be matched exactly, those clusters are likely to appear as imbalanced in the minibatches, especially if the size of the minibatch is small and does not respect the statistics of the original distribution.  Due to the marginal constraints, samples from one cluster are likely to be matched to unrelated clusters, as depicted in Figure~\ref{fig:mbot_non_opt_connection}. This explains why in practice previous works relied on large minibatches to mitigate this sampling effect~\cite{deepjdot}.  To overcome this issue, we propose the natural solution of relaxing the marginal constraints at the minibatch level. The expected outcome is twofold: {\em i)} mitigating the effect of subsampling in the minibatch strategy and {\em ii)} providing a natural and scalable robust optimal transport computation strategy at the global level. We discuss in the following some theoretical considerations to support this claim. 
}
%
%
%
%

\paragraph{Theoretical analysis: impact of an outlier} 
{We start by examining the impact of an outlier in the behaviors of OT and UOT. The following lemma illustrates the relations between those two quantities.}

\begin{lemma}\label{lemma:robust}
	Take $(\alpha,\beta)$ two probability distributions. For $\zeta\in[0,1]$, write $\tilde{\alpha} = \zeta\alpha + (1-\zeta)\delta_\zz$ a distribution  perturbed by a Dirac outlier located at some $z$ outside of the support of $(\alpha,\beta)$. Take the unregularized OT loss $\operatorname{OT}_{KL}^{\tau, 0}$ with KL entropy and cost $C$. Write $m(\zz) =\int  C(\zz,\yy) d\beta(\yy)$. One has:
	\begin{align}\label{lemma:eq-bound-uot-outlier}
		\operatorname{OT}_\texttt{KL}^{\tau, 0}(\tilde{\alpha}, \beta, C) &\leq \zeta \operatorname{OT}_\texttt{KL}^{\tau, 0}(\alpha, \beta,C)\\
		& + 2\tau(1-\zeta)(1 - e^{-m(\zz) / 2\tau})\nonumber
	\end{align}
    Now take the unregularized, balanced OT loss $\operatorname{OT}_\phi^{\infty, 0}$ with cost $C$. Write $(f,g)$ the optimal dual potentials (i.e. functions) of $\operatorname{OT}_\phi^{\infty, 0}(\alpha,\beta)$, and $y^*$ in $\beta$'s support. Then:
	\begin{align}\label{lemma:eq-bound-ot-outlier}
	\operatorname{OT}_\phi^{\infty, 0}(\tilde{\alpha}, \beta) &\geq \zeta \operatorname{OT}_\phi^{\infty, 0}(\alpha, \beta)  \\
    &+ (1-\zeta)\Big(C(\zz, y^*) - g(y^*) + \int gd\beta\Big)\nonumber
	\end{align}
\end{lemma}

Equation~\eqref{lemma:eq-bound-ot-outlier} shows that when $\zz$ gets further from the supports of $(\alpha,\beta)$, the OT loss increases. 
However for UOT the upper bound~\eqref{lemma:eq-bound-uot-outlier} tends to saturate as $\zz$ gets further away. What remains is the UOT loss between distributions whose outliers are removed, with a cost of removing the outlier proportional to its mass. 


{
We first illustrate Lemma \ref{lemma:robust} with a toy example in Figure \ref{fig:deviation}. We consider a probability distribution $\alpha$  tainted with an outlier (green dot) to get a target probability distribution $\alpha^\prime = \frac{1}{n+1}(n.\alpha + \delta_\zz)$. We then move away the outlier from $\alpha$'s support, as shown with the green arrow, and we calculate several OT costs. The minibatch size is set to $m=5$ and the total cost is the average of $k$ OT costs between those minibatches. We see that OT variants are not robust to the outlier as their loss increases along the outlier displacement unlike UOT variants which reach a plateau as predicted by Lemma \ref{lemma:robust}. Each computation is done 5 times to show that variance is lower for bigger $k$ and that UMBOT has a lower variance than MBOT for $k=30$ and $k=500$.

We consider now an example in 2D, akin to Figure~\ref{fig:mbot_non_opt_connection}, where our goal is to illustrate the OT plan between two empirical distributions of 10 samples in Figure \ref{fig:2D_gauss}. We use two 2D empirical distributions  where the samples belong to a certain cluster depending on a related class (color information). The source data are equally distributed between classes while the target data have different proportions, 3 samples belong to the red class while 7 samples belong to the green class. Different proportions between domains are ubiquitous for real world data. We compare unbalanced minibatch OT, minibatch OT, entropic OT and UOT. For UOT, the divergence $D_\phi$ equals to \texttt{KL} divergence and for the minibatch variant, the minibatch size is $m=2$. We can see from the OT plans in the first row of the figure that the cluster structure is more or less recovered. However OT and minibatch OT tend to connect samples from different classes. This configuration would lead, for instance, to negative transfer in a context of domain adaptation applications, \emph{i.e.,} matching of samples between different domains. This is less true for UOT, where the pairings between different classes is diminished and  tend to disappear when we reduce the penalty $\tau$. 
 \begin{figure*}[h]
     \centering
     \includegraphics[scale=0.35]{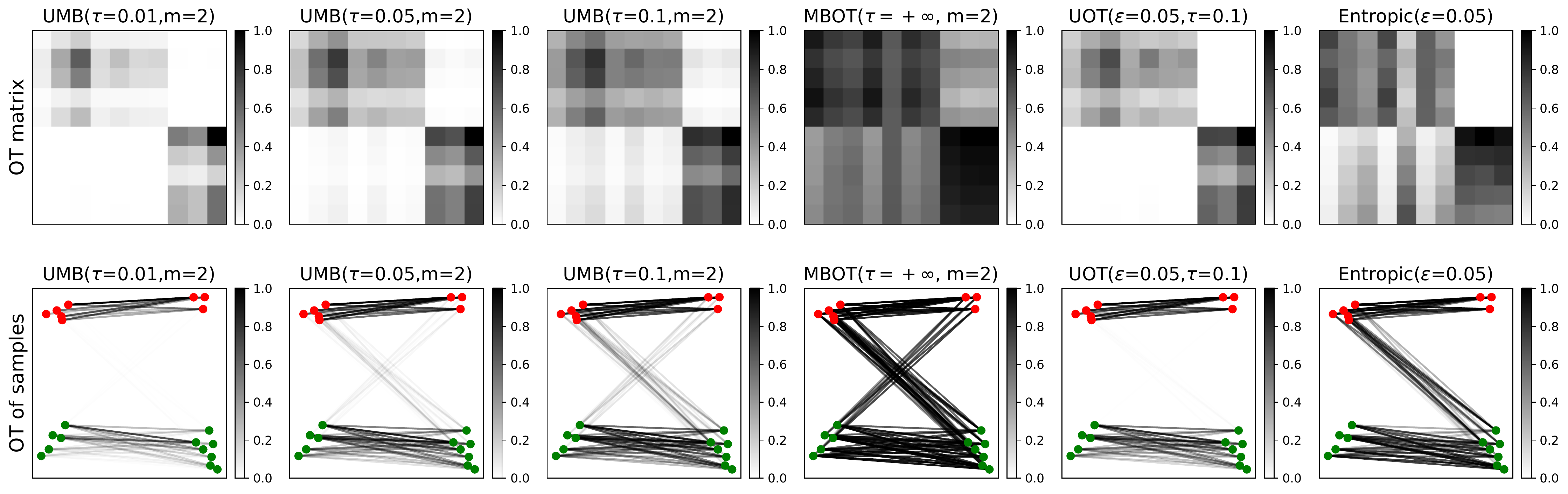}
     \vspace{-0.3cm}
     \caption{ Several OT plans, normalized by their maximum value, between 2D distributions with $n=10$ samples. The first row shows the minibatch OT plans $\overline{\Pi}^m$ for different values of $m$, {the second row provides an equivalent geometric interpretation of the OT plans, where the mass transportation is depicted as connexions between samples.}\label{fig:2D_gauss}}
 \end{figure*}
}

\section{Unbalanced Minibatch Optimal Transport}\label{sec:mbuot}

In this section we express some mathematical properties at the heart of this
work. In \cite{fatras20a},  authors described some properties of
the minibatch OT. We provide here extensions of those results to Unbalanced
OT. We start by defining
minibatch estimators, then we review the concentration bounds and finish with
optimization properties. To derive concentration bounds we first prove 
that the UOT cost is finite and the optimal transport plan is bounded. Without loss of generality, we consider $n$-tuples $\XX$ and $\YY$ with uniform vectors $\uu_n$,
to form empirical distributions encountered in the different applications and the associated ground cost matrix $C$.

\subsection{Minibatch Unbalanced OT estimation}

{\bfseries Estimators.}
To build minibatches, we select $m$ samples from $\XX$ and $\YY$. We rely on a generic element of indices $I = (i_1,\ldots,i_m) \in \llbracket n \rrbracket^m$, which is called an index $m$-tuple. $I$ represents the selected samples from the $n$-data tuple $\XX$ or $\YY$. In this work, we only focus on $m$-tuples without replacement $I$, whose their set is denoted $\PP$.

\begin{definition}[Minibatch UOT]\label{def:minibatch_wasserstein} Let $C = C^n(\XX, \YY)$ be a square matrix of size $n$. Given an unbalanced OT loss $h \in \{\operatorname{OT}_\phi^{\tau, \varepsilon}, S_\phi^{\tau, \varepsilon}\}$ and an integer $m \leq n$, we define the following quantity:
\begin{equation}
\bar{h}_C^m(\XX, \YY) :=  \frac{(n-m)!^2}{n!^2} \sum_{I,J \in \PP} h(\uu_m, \uu_m, C_{I, J})
\label{def:discrete}
\end{equation}
where for $I,J$ two $m$-tuples, $C_{(I,J)}$ is the matrix extracted from $C$ by keeping the rows and columns corresponding to $I$ and $J$ respectively. We denote the optimal plan $\Pi_{ I, J }$, lifted as a $n \times n$ matrix where all entries are zero except those indexed in $I \times J$. We define the \textit{averaged minibatch transport plan}:
 $\overline{\Pi}^m(\XX, \YY)  \defas \dbinom{n}{m}^{-2} \sum_{I,J \in \PP} \Pi_{  I, J }.$
\end{definition}

We omit $C$ when clear from context. A simple combinatorial argument provided in
appendix assures that the sum  of $\uu_m$ over all
$m$-tuples $I$ gives $\uu_n$. 
In the formulation above, we no longer compute UOT between the full distributions
but instead we compute the expectation of UOT over all minibatches
drawn from $\alpha^{\otimes m } \otimes \beta ^{\otimes m}$:
\begin{equation}
  \hspace{-0.2cm}
  E_h \defas  \underset{(\XX,\YY) \sim  \alpha^{\otimes m } \otimes \beta ^{\otimes m}}{\mathbb{E}} [h(\uu_m, \uu_m, C^m(\XX, \YY))],
  \label{def:expectationMinibatch}
\end{equation}

The combinatorial number of terms is prohibitive to compute, fortunately we can rely on subsample quantities.

\begin{definition}[Minibatch subsampling]\label{def:sub_discrete} Consider the notations of definition \ref{def:minibatch_wasserstein}. Pick an integer $ k > 0$, we define:
  \begin{equation}
    \widetilde{h}_{k,C}^m(\XX, \YY) :=k^{-1} \sum_{  (I, J)  \in D_k  } h(\uu_m, \uu_m, C_{I, J})
  \end{equation}
where $D_k$ is a set of cardinality $k$ whose elements are drawn at random from
the uniform distribution on $ \Gamma:= \mathcal{P}_m \times \mathcal{P}_m  $. A
similar construction holds for incomplete minibatch transport plan denoted
as $\widetilde{\Pi}_k^m(\XX, \YY)$.
\end{definition}

Note that $\bar{h}^m$ and $\widetilde{h}_k^m$ are unbiased estimators of $E_h$
as they are, respectively, complete and incomplete U-statistics
\cite{LeeUstats}.  The minibatch UOT losses are positive and symmetric, however they are not definites, \emph{i.e.,} $\bar{h}^m(\XX, \XX) > 0$ for non trivial $\XX$ and $1 < m < n$.


\subsection{Deviation bounds}\label{sec:asymp_conv}
{
{Our first lemma intends to show that the UOT cost is finite and that the optimal transport plan is bounded, which is needed to establish the concentration bounds.}
\begin{lemma}[Bounded UOT and optimal transport plan]\label{app:bounded_plan}
	Let $C$ be a ground cost and $\aa, \bb$ two positive vectors in $\R_+^n$ such
	that $m_\aa=\|\aa\|_1>0$ and  $m_\bb =\|\bb\|_1>0$. Assume that $\langle
	\aa\bb^\top, C \rangle < +\infty$. Consider $h = \operatorname{OT}_\phi^{\tau,
	\varepsilon}$ and assume $\varepsilon > 0$ or $\phi^\prime_\infty >0$. Then
	$h(\aa,\bb, C)$ is finite and the set of optimal transport plan is a compact set.
\end{lemma}

It is straightforward to prove boundedness of $h = S_\phi^{\tau, \varepsilon}$
from Lemma \ref{app:bounded_plan}. {We can now turn to establish} concentration bounds for both
incomplete estimators $\widetilde{h}_k^m$ and $\widetilde{\Pi}_k^m$.}

\begin{theorem}[Maximal deviation bound]\label{thm:inc_U_to_mean} Let $ \delta \in (0,1) $, three integers $k \geq 1$ and $m \leq n$ be fixed. Consider two $n$-tuples $\XX \sim \alpha^{\otimes n}$ and $\YY \sim \beta^{\otimes n}$ and a kernel $h\in \{\operatorname{OT}_\phi^{\tau, \varepsilon}, S_\phi^{\tau, \varepsilon} \}$. We have a maximal deviation bound between $\widetilde{h}_k^m(\XX, \YY)$ and $E_h$ depending on the number of samples $n$ and the number of batches $k$. With probability at least $1-\delta$ on the draw of $\XX, \YY$ and $D_k$ we have:
  \begin{align*}
    \vert \widetilde{h}_k^m(\XX, \YY) - E_h \vert \leq M \left(\sqrt{\frac{ \log(\frac{2}{\delta})}{2\lfloor \frac{n}{m} \rfloor}} + \sqrt{\frac{2\log(\frac{2}{\delta})}{k}  }\right),
  \end{align*}
where {$M$ is the UOT upper bound.} Furthermore, for $h = \operatorname{OT}_\phi^{\tau, \varepsilon}$, let $\mathfrak{M}_\Pi^\infty$ be the maximum mass of minibatch transport plans. For all $ k \geqslant 1 $, all $ 1 \leqslant i \leqslant n  $, with probability at least $1-\delta$ on the draw of $\XX, \YY$ and $D_k$ we have:
\begin{equation*}
\vert  \widetilde{\Pi}_k^m(\XX, \YY)_{(i)} \mathbf{1}_n - \overline{\Pi}^m(\XX, \YY)_{(i)} \mathbf{1}_n \vert \leqslant \mathfrak{M}_\Pi^\infty \sqrt{\frac{2 \log(2/\delta)}{k}},
\end{equation*}
where we denote by $\Pi_{(i)}$ the $i$-th row of matrix $\Pi$ and by
$ \mathbf{1}_n \in \R^n $ the vector whose entries are all equal to $1$.
\end{theorem}

This deviation bound shows that if we increase the number of data $n$ and batches $k$ while keeping the minibatch size $m$ fixed, we get closer to the expectation. The rate $\frac{m}{n}$ is almost optimal and is the same as in \cite{fatras20a}. The main difference is the upper bound $M$ which bounds UOT. Note that the bound does not depend on the dimension of $\mathcal{X}$ unlike original unbalanced OT \citep{sejourne2019sinkhorn}. Regarding OT plans, the constant $\mathfrak{M}_\Pi$ represents the maximum mass of minibatch transport plans which would be 1 for OT. 

\long\def\comment#1{Another interesting output of minibatch UOT is the average matrix $\overline{\Pi}^m$. We investigate the deviation of its marginals and those from its incomplete estimators $\widetilde{\Pi}_k^m$. In what follows, we denote by $\Pi_{(i)}$ the $i$-th row of matrix $\Pi$ and by
$ \mathbf{1}_n \in \R^n $ the vector whose entries are all equal to $1$.

\begin{theorem}[Distance to marginals]\label{thm:dist_marg} Let $ \delta \in (0,1) $, two integers $m \leq n$ be fixed. Consider two $n$-tuples $\XX \sim \alpha^{\otimes n}$ and $\YY \sim \beta^{\otimes n}$, $h = \operatorname{OT}_\phi^{\tau, \varepsilon}$ and $\mathfrak{M}_\Pi$ be the maximum mass of minibatch transport plans. For all $ k \geqslant 1 $, all $ 1 \leqslant i \leqslant n  $, with probability at least $1-\delta$ on the draw of $\XX, \YY$ and $D_k$ we have:
\begin{equation*}
\vert  \widetilde{\Pi}_k^m(\XX, \YY)_{(i)} \mathbf{1}_n - \overline{\Pi}^m(\XX, \YY)_{(i)} \mathbf{1}_n \vert \leqslant \mathfrak{M}_\Pi \sqrt{\frac{2 \log(2/\delta)}{k}}.
\end{equation*}
\end{theorem}

The deviation linearly decreases as a the number of minibatches $k$ grows. The constant $\mathfrak{M}_\Pi$ represents the mass of transport plan which would be 1 for OT. This can be especially important for applications which need the transport plan such as color transfer, see \cite{fatras20a}.} 


\begin{figure*}[t]
    \centering
    \includegraphics[scale=0.45]{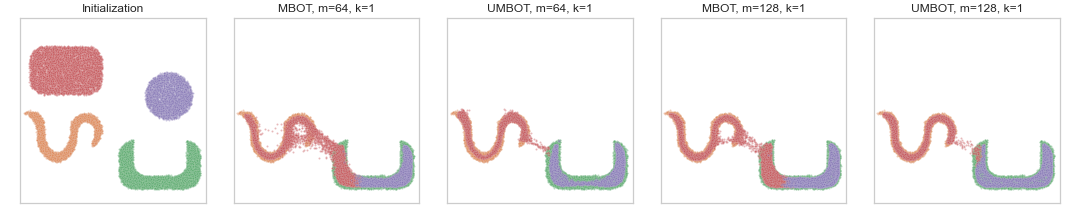}
    \vspace{-0.6cm}
    \caption{(Best viewed in colors) Minibatch UOT gradient flow on a 2D dataset. Source data and target data are divided in 2 imbalanced clusters, source left (red) and target right (green) shapes have 6400 samples while source right (purple) and target left (orange) shapes have 3600 samples. The batch size $m$ is set to $\{64,128\}$ and the number of minibatch $k$ is set to 1, meaning that the explicit Euler integration step is conducted for each batch. Results are computed with the (unbalanced) minibatch Sinkhorn divergence losses.}
    \label{fig:GF_2D}
\end{figure*}

\subsection{Unbiased gradients and optimization}

The Wasserstein distance is known to suffer from biased gradients
\citep{Bellemare_cramerGAN}, meaning that minimizing the estimator of the
Wasserstein distance with empiricial distributions does not lead to the minimum
of the Wasserstein distance between full distribution. While
 minibatch entropic OT does not suffer from these biased gradients~\cite{fatras20a}, 
in this section we show that this property remains true for minibatch
UOT, including unregularized UOT. We achieve this point by relying on Clarke regularity.

We study a standard parametric data fitting problem. Given some discrete samples \(\left(x_{i}\right)_{i=1}^{n} \subset \mathcal{X}\) from an unknown distribution \(\alpha\), our goal is to fit a parametric model \(\lambda \mapsto \beta_{\lambda} \in \mathcal{M}_+(\mathcal{X})\) to \(\alpha\) for $\lambda \in \Lambda$ in an Euclidian space. To do so, we use minibatch UOT and its incomplete estimators $\widetilde{h}_k^m$ as a contrast loss. U-statistics as contrast loss have been studied in \cite{papa_NIPS2015}. Stochastic gradients (SGD) has also proven to be really efficient at optimizing neural network parameters $\lambda$ even if they are non convex \cite{Bottou10}. To justify the convergence of SGD, we need to exchange expectation and gradients and use that $\widetilde{h}_k^m$ is an unbiased estimator of $E_h$. 

The former is not immediate because UOT is not differentiable as we do not have a unique optimal transport plan when $\varepsilon=0$. Thus, we use the notion of Clarke generalized derivatives. They define a regularity for nonsmooth but locally Lipschitz and semi-continuous function. It is a close concept to subgradients for convex functions since when a convex function is locally Lipshitz at $\xx$ the two notions are equivalent. An intuitive geometric interpretation is that a function is Clarke regular if it doesn't have "upwards dashes" in its graph, for a total survey see \citep{clarke1990optimization}. 

\begin{theorem}\label{thm:exchange_grad_exp_sm_app}
Let $\hat{\XX}, \{\hat{\YY}_\theta\}_{\theta \in \Theta}$ be two $m$-tuples of random vectors compactly supported, $h \in \{\operatorname{OT}^{\tau, \varepsilon}, S^{\tau, \varepsilon} \}$ and $C^m$ a $\mathbf{C}^1$ cost.
Under an additional integrability assumption, we have:
\begin{equation*}\label{eq:exchange_theorem_eq2_proof}
\partial_{\mathbf{\theta}} \expect [ h(\uu, \uu, C^m(\hat{\XX}, \hat{\YY}_{\theta})) ] =  \expect [ \partial_{\mathbf{\theta}} h(\uu, \uu, C^m(\hat{\XX}, \hat{\YY}_{\theta})) ],
\end{equation*}
with both expectation being finite. Furthermore the function $\theta \mapsto - \expect [ h(\uu, \uu, C^m(\hat{\XX}, \hat{\YY}_{\theta})) ]$ is also Clarke regular.
\end{theorem}

Note that Theorem \ref{thm:exchange_grad_exp_sm_app} implies that if we use the
Minibatch UOT loss with $h \in \{\operatorname{OT}_\phi^{\tau, \varepsilon},
S_\phi^{\tau, \varepsilon}\}$  as a loss function, then the minus objective
function is Clarke regular. Furthermore, Stochastic gradient with decreasing step sizes converges almost surely to the set of critical points of Clarke generalized derivative \citep{davis2020stochastic, majewski2018analysis}.
 As a consequence, it is justified to use SGD with minibatch UOT, as it converges to the optimal $\lambda^*$.

\section{Experiments}
In this section, we illustrate the practical behavior of unbalanced minibatch OT for gradient flow and
for domain adaptation experiments. We relied on the POT
package \cite{flamary2017pot} to compute the exact OT solver or the entropic UOT
loss and the Geomloss package \cite{feydy19a} for the Unbalanced Sinkhorn
divergence. The experiments were designed in PyTorch \cite{paszke2017automatic}
and all the code will be released upon publication.

\subsection{Unbalanced MiniBatch OT gradient flow}

Consider a given target distribution $\alpha$, the purpose of gradient flows is to model a distribution $\beta(t)$ which at each iteration follows the gradient direction minimizing the loss $\beta_t \mapsto h(\alpha, \beta_t)$  \cite{Peyre2015, liutkus19a}. The gradient flow simulate the non parametric setting of data fitting problem, where the modeled distribution $\beta$ is parametrized by a vector position $\xx$ that encodes its support.

We follow the same experimental procedure as in \cite{feydy19a}. The gradient flow algorithm uses a simple explicit Euler integration scheme. Formally, it starts from an initial distribution at time $t=0$ and integrates at each iteration a SDE.
In our case, we cannot compute the gradient directly from our minibatch OT losses. As the OT loss inputs are empirical distributions, we have an inherent bias when we calculate the gradient from the weights $\frac{1}{m}$ of samples that we correct by multiplying the gradient by the inverse weight $m$. Finally, we integrate: $\dot{\XX}(t)=-m \nabla_{\boldsymbol{\XX}}\widetilde{h}_k^m(\XX, \YY)$.

For $\alpha$ and $\beta(0)$ we generate 10000 2D points divided in 2 imbalanced
clusters with number of samples in each cluster provided in Figure \ref{fig:GF_2D}. We consider the
(unbalanced) sinkhorn divergence, a squared euclidean cost, a learning rate of 0.02, 5000 iterations,
$m$ equals 64 or 128 and $k=1$. We show the gradient flow of the upper clusters to the lower clusters
in Figure \ref{fig:GF_2D}. 
From the experiment, we can see that the minibatch OT is not robust to
imbalanced classes on the contrary to the minibatch UOT. Indeed there are data
from the upper left cluster which converge to the down right cluster and we can
also see an overlap between the classes. Due to OT marginal constraints, the
loss forces to transport all data in the batch which results in breaking the
target shapes. This is not the case for minibatch UOT, which better respects the
shape of target distributions.

\long\def\comment#1{
{\bfseries Bias}
From the Sinkhorn divergence experiments, we can see that the final distribution is not the original target distribution but a shrunk version of it. For a large enough batch size, one can see that we are able to catch the complex geometry of the target distribution but can not recover the original distribution. We catch the empty slots and the curved tail. As the batch size becomes small, we loose the complex geometry and the final distribution lives inside the denser area.

This behavior is due to the bias from the minibatch OT losses. The bias has the effect to favor dense area over low dense area. As the batch distribution follows the data distribution, it is more likely to pick data from the dense regions inside the minibatch than from a low dense area or the border of the distribution. That's why the gradient flow sends the low dense area toward the denser area and does not respect the symmetry in the target distribution. We can see it the curved tail which does not fit the target samples because they are driven to the denser area. It explains why one should not take a too small batch size. This behavior is stressed by increasing the number of averaged OT terms. One can see that it helps to leave the low dense area inside the dense area but do not change the full behavior.

{\bfseries Convergence and gap}
We also ploted the loss value $S_{\varepsilon}(\alpha, \beta_t)$ according to the execution time. The experiments were processed on an \textit{1,8 GHz Intel Core i5} processor. We observed that the final value is not zero and increases when the batch size gets smaller. This is expected as we loose a lot of information about the low dense area. However, we can see a convergence in the loss value. The final value is more stable when $K$ increases but the behavior is quite comparable. Indeed increasing $K$ does not lead to a smaller loss value between the final distribution and the target distribution. that's why if one wants to have a final distribution closer to the true OT losses, one should not increase $K$ but the should increase the batch size instead.

}


\begin{table*}[t!]
    \center
    \small
  \begin{tabular}{|@{\hskip3pt}c@{\hskip3pt}|@{\hskip3pt}c@{\hskip3pt}|c|c|c|c|c|c|c|c|c|c|c|c|c|}
    \hline
     \multirow{8}{*}{\textsc{da}} & Method & A-C & A-P & A-R & C-A & C-P & C-R & P-A & P-C & P-R & R-A & R-C & R-P & avg\\
     \hline
     & \textsc{resnet}-50 & 34.9 & 50.0 & 58.0 & 37.4 & 41.9 & 46.2 & 38.5 & 31.2 & 60.4 & 53.9 & 41.2&  59.9 & 46.1\\
     & \textsc{dann} (*)  & 44.3 & 59.8 & 69.8 & 48.0 & 58.3 & 63.0 & 49.7 & 42.7 & 70.6 & 64.0 & 51.7 & 78.3 & 58.3 \\
     & \textsc{cdan-e}(*) & 52.5 & 71.4 & 76.1 & 59.7 & 69.9 & 71.5 & 58.7 & 50.3 & 77.5 & 70.5 & 57.9 & \textbf{83.5} & 66.6 \\
     & \textsc{deepjdot} (*) & 50.7 & 68.6 & 74.4 & 59.9 & 65.8 & 68.1 & 55.2 & 46.3 & 73.8 & 66.0 & 54.9 & 78.3 & 63.5\\
     & \textsc{alda} (*) & 52.2 & 69.3 & 76.4 & 58.7 & 68.2 & 71.1 & 57.4 & 49.6 & 76.8 & 70.6 & 57.3 & 82.5 & 65.8\\
     & \textsc{rot} (*) & 47.2 & 71.8 & 76.4 & 58.6 & 68.1 & 70.2 & 56.5 & 45.0 & 75.8 & 69.4 & 52.1 & 80.6 & 64.3 \\
     & \textsc{jumbot} & \textbf{55.2} & \textbf{75.5} & \textbf{80.8} & \textbf{65.5}  & \textbf{74.4} &  \textbf{74.9}  & \textbf{65.2}  & \textbf{52.7} & \textbf{79.2} & \textbf{73.0} & \textbf{59.9} & 83.4 & \textbf{70.0}\\
     \hline
     \hline
     \multirow{6}{*}{\textsc{pda}} & \textsc{resnet-50} & 46.3 & 67.5 & 75.9 & 59.1 & 59.9 & 62.7 & 58.2 & 41.8 & 74.9 & 67.4 & 48.2 & 74.2 & 61.4\\
     & \textsc{deepjdot}(*) & 48.2 & 66.2 & 76.6 & 56.1 & 57.8 & 64.5 & 58.3 & 42.7 & 73.5 & 65.7 & 48.2 & 73.7 & 60.9 \\
     & \textsc{pada} & 51.9 & 67.0 & 78.7 & 52.2 & 53.8 & 59.0 & 52.6 & 43.2 & 78.8 & 73.7 & 56.6 & 77.1 & 62.1\\
     & \textsc{etn} & 59.2 & 77.0 & 79.5 & 62.9 & 65.7 & 75.0 & 68.3 & 55.4 & 84.4 & 75.7 & 57.7 & \textbf{84.5} & 70.4\\
     & \textsc{ba3us}(*) & 56.7 & 76.0 & \textbf{84.8} & 73.9 & 67.8 & \textbf{83.7} & 72.7 & 56.5 & 84.9 & 77.8 & 64.5 & 83.8 & 73.6 \\
     & \textsc{jumbot} & \textbf{62.7} & \textbf{77.5} & 84.4 & \textbf{76.0} & \textbf{73.3} & 80.5 & \textbf{74.7} & \textbf{60.8} & \textbf{85.1} & \textbf{80.2} & \textbf{66.5} & 83.9 & \textbf{75.5}\\
     \hline
  \end{tabular}
  \vspace{-0.2cm}
  \caption{Summary table of DA and Partial DA results on Office-Home (ResNet-50). (*) denotes the reproduced methods.}
  \label{tab:office_home}
\end{table*}

\subsection{Domain adaptation}
 We now follow the settings of unsupervised DA where both domains share the same labels $\mathcal{Y}_s=\mathcal{Y}_t$.
 
\textbf{\textsc{jumbot}.} Optimal Transport has been proposed in \cite{Courty2017} as
a way to solve the domain adaptation problem. Our method is based on \cite{deepjdot} and aims at finding a joint distribution map between a source and a target distribution by taking into account a term on
a neural network embedding space and on the label space. Formally, let
$g_\theta$  be an embedding function where the input is mapped into the latent
space and $f_\lambda$ which maps the latent space to the label space on the
target domain. The embedding space is in our experiment the before last layer of
a neural network. For a given minibatch, embedding $g_\theta$ and classification
map $f_\lambda$, the transfer term is:
{
\begin{align}\label{def:jdot}
  &\qquad \bar{h}_{C_{\theta, \lambda}}^m((\XX^s,\YY^s),(\XX^t,f_\lambda(g_\theta(\XX^t)))), \text{with } \\
  &  (C_{\theta, \lambda})_{i,j}=\eta_1 \|g_\theta(\xx_i^s) - g_\theta(\xx_j^t)\|_2^2 + \eta_2 \mathcal{L}(\yy_i^s, f_\lambda(g_\theta(\xx_j^t))) \nonumber,
\end{align}}
where $\mathcal{L}(.,.)$ is the cross-entropy loss and $\eta_1, \eta_2$ are positive constants. Basically, this specific transportation cost combines a distance between the representation of the data through the neural network $g_\theta$ and a loss function between the associated labels~\cite{Courty2017}. Taking $k=1$ led to state of the art results. The Csisz\`ar divergence $\phi$ is the Kullback-Leibler divergence $\texttt{KL}$. We also add a cross entropy term on the source data. Hence our optimization problem is:
\begin{align}
   \underset{\theta, \lambda}{\operatorname{min}} &\sum_i \mathcal{L}(f_\lambda(g_\theta(\xx_i^s)), \yy_i^s)  \\
   & + \eta_3 \widetilde{h}_{k,C_{\theta, \lambda}}^m((\XX^s,\YY^s),(\XX^t,f_\lambda(g_\theta(\XX^t)))). \nonumber 
\end{align}
Our method is called \textbf{\textsc{jumbot}} and stands for Joint Unbalanced MiniBatch OT. It is related to \textsc{deepjdot} \cite{Courty2017, deepjdot} at the notable exceptions that we use minibatch UOT, which can also handle partial domain adaptation as suggested by our experiments. 

{\bfseries Datasets.} We start with \textbf{digits} datasets. Following the evaluation protocol of \cite{deepjdot} we experiment on three adaptation scenarios: USPS
to MNIST (U$\mapsto$M), MNIST to M-MNIST (M$\mapsto$MM), and SVHN
to MNIST (S$\mapsto$M). MNIST \cite{MNIST} contains 60,000
images of handwritten digits, M-MNIST contains the 60,000 MNIST
images with color patches \cite{DANN} and USPS \cite{USPS} contains
7,291 images. Street View House Numbers (SVHN)\cite{svhn} consists of 73, 257 images with digits and numbers in natural scenes. We report the evaluation results on the test target datasets.
\textbf{Office-Home} \cite{office_home} is a difficult dataset for unsupervised domain adaptation (UDA), it has 15,500 images from four different domains: Artistic images
(A), Clip Art (C), Product images (P) and Real-World
(R). For each domain, the dataset contains images of 65
object categories that are common in office and home scenarios. We evaluate all methods in 12 adaptation scenarios.
\textbf{VisDA-2017} \cite{visda} is a large-scale dataset for UDA from simulation to real. VisDA contains 152,397 synthetic images as the source domain and 55,388 real-world images as the target domain. 12 object categories are shared by these
two domains. Following \cite{cdan2018, alda2020}, we evaluate all methods on VisDA validation set.
\begin{table}[t]
\small
        \begin{center}
            \begin{tabular}{|@{\hskip3pt}c@{\hskip3pt}|@{\hskip3pt}c@{\hskip3pt}|@{\hskip3pt}c@{\hskip3pt}|@{\hskip3pt}c@{\hskip3pt}|@{\hskip3pt}c@{\hskip3pt}|}
                 \hline
                 Methods/DA & U $\mapsto$ M & M $\mapsto$ MM & S $\mapsto$ M & Avg\\
                 \hline
                  \textsc{dann}(*) & 92.2 $\pm$ 0.3 & 96.1 $\pm$ 0.6 & 88.7 $\pm$ 1.2 & 92.3\\
                 \textsc{cdan-e}(*) & \textbf{99.2 $\pm$ 0.1} &  95.0 $\pm$ 3.4 & 90.9 $\pm$ 4.8 & 95.0 \\
                 \textsc{alda}(*) & 97.0 $\pm$ 1.4 & 96.4 $\pm$ 0.3 & 96.1 $\pm$ 0.1 & 96.5\\
                 \textsc{deepjdot}(*) & 96.4 $\pm$ 0.3  & 91.8 $\pm$ 0.2  & 95.4 $\pm$ 0.1 & 94.5 \\
                 \textsc{jumbot} & 98.2 $\pm$ 0.1  & \textbf{97.0 $\pm$ 0.3}  & \textbf{98.9 $\pm$ 0.1} & \textbf{98.0} \\
                 \hline
            \end{tabular}
        \end{center}
    \vspace{-0.1cm}
    \caption{Summary table of DA results on digit datasets. Experiments were run three times. (*) denotes the reproduced methods.}
    \label{tab:digits_results}
\end{table}

{\bfseries Results.} We compare our method against state of the art methods: \textsc{dann}\cite{DANN}, \textsc{cdan-e}\cite{cdan2018}, \textsc{deepjdot}\cite{deepjdot}, \textsc{alda}\cite{alda2020}, \textsc{rot}\cite{balaji2020robust}. Details about training procedure and architectures can be found in supplementary.

{We reported the test score at the end of optimization for all benchmarks and methods.} The results on digit datasets can be found in Table \ref{tab:digits_results}
where (*) denotes reproduced results. We conducted each experiment three
times and report their average results and variance. For fair comparisons, we only resize and
normalize the image without data augmentation. We see that our method
performs  best with a margin of at least $1.5$ points. Furthermore, we see an important $4\%$ increase of the performance compared to \textsc{deepjdot}. A deeper
analysis of this difference is considered in the next paragraph.
Office-Home results are gathered in Table \ref{tab:office_home} and VisDA are
reported in Table \ref{tab:summary_visda}. For fair comparison with previous
work, we used a similar data pre-processing and we used the ten-crop technique
\cite{cdan2018, alda2020} for testing our methods. Experiments in
\cite{balaji2020robust} follow a different setup explaining the difference of performance between
their reported score and our reproduced score. \textsc{jumbot} achieves the
best accuracy on average and on 11 of the 12 scenarios on the Office-Home
dataset and achieves the best accuracy on VisDA at the end of training with a margin of 4\% and 2\% respectively.

\begin{figure}[t!]
    \includegraphics[width=\linewidth,height=4cm]{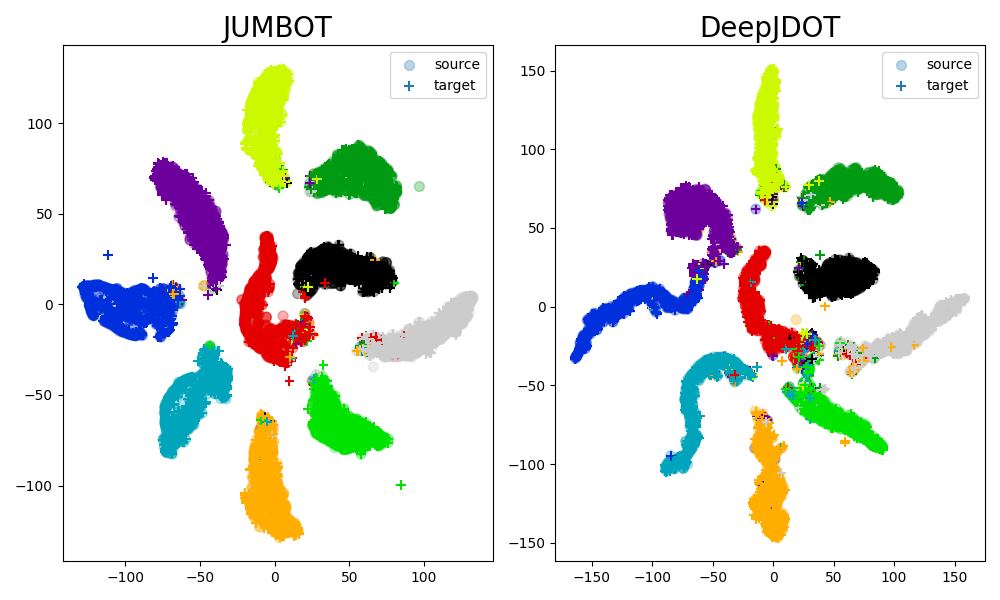}
    \vspace{-0.7cm}
    \caption{T-SNE embeddings of 10000 test samples for MNIST (source) and MNIST-M(target) for \textsc{deepjdot} and our method. It shows the  ability  of  the  methods  to  discriminate  classes  (samples  are  colored  w.r.t.  their classes).}
    \label{fig:tsne}
\end{figure}

\begin{table}[t]
        \begin{center}\small
            \begin{tabular}{|@{\hskip3pt}c@{\hskip3pt}|@{\hskip3pt}c@{\hskip3pt}|}
                 \hline
                 Methods & Accuracy (in $\%$)\\
                 \hline
                 \textsc{cdan-e}(*) & 70.1 \\
                 \textsc{alda}(*) & 70.5\\
                 \textsc{deepjdot}(*) & 68.0 \\
                 \textsc{robust ot}(*) & 66.3 \\
                 \textsc{jumbot} & \textbf{72.5} \\ 
                 \hline
            \end{tabular}
        \end{center}
     \vspace{-0.2cm}
    \caption{Summary table of DA results on VisDA datasets. (*) denotes the reproduced methods.}
    \label{tab:summary_visda}
\end{table}

{\bfseries Analysis.} In this paragraph, we study the difference of behavior
between \textsc{deepjdot} and \textsc{jumbot}. Along \textsc{jumbot}'s training on
the DA task USPS to MNIST, we measured the percentage of mass between data with
different label at each iteration. In average along training about $7\%$ of
\textsc{deepjdot} connections are between data with different labels while this
percentage decreases to $0.7\%$ for \textsc{jumbot}. So \textsc{deepjdot} transfers
wrong labels to the target which will decrease the overall accuracy. We also plot a TSNE embedding
of our method and \textsc{deepjdot} (see Figure \ref{fig:tsne}), we can see that
there are some overlaps between clusters for \textsc{deepjdot} unlike our method. This is probably due to the minibatch smoothing effect which would tend to bring clusters of different classes closer. We
provide in appendix a classification accuracy along training which demonstrates
the network overfitting with \textsc{deepjdot} and not with \textsc{jumbot}.
Finally, we also provide in supplementary a sensitivity analysis to the parameters, showing
that \textsc{jumbot} is more robust to small batch size than \textsc{deepjdot} which
is interesting for small computation budget.



\subsection{Partial DA}

Finally we consider the Partial DA (PDA) application. In PDA, the target labels are a subset of the source labels, \emph{i.e.,
$\mathcal{Y}_t \subset \mathcal{Y}_s$}. Samples belonging to these missing
classes become outliers which can produce negative transfer. We want to investigate
the robustness of our method in such an extreme scenario. We evaluate our method
on partial Office-Home, where we follow \cite{Cao_2018_ECCV} to
select
the first 25 categories (in alphabetic order) in each domain as a partial target
domain. We compare our method against state of the art PDA methods:
\textsc{pada} \cite{Cao_2018_ECCV}, \textsc{etn} \cite{ETN_2019_CVPR} and
\textsc{ba3us} \cite{liang2020baus}. For fair comparison we followed the
experimental setting of \textsc{pada}, \textsc{etn} and \textsc{ba3us} and we
report to the supplementary material for more details. {The final performances are gathered
in the lower part of table \ref{tab:office_home}. Note that we do not
use the ten-crop technics for evaluating the methods, as we were not able to reproduce \textsc{pada} 
and \textsc{etn}.} We can see that \textsc{jumbot} is state of the art on 9 of the
12 domain adaptation tasks and is on average 2\% above competitors. Finally, we also 
evaluate \textsc{deepjdot} \cite{deepjdot} on
the PDA task, \textsc{jumbot} is on average 15\% higher on this problem despite the strong similar nature of the way the problem is
solved, showing the clear advantages of our strategy.


\section{Conclusion}

Computing minibatches is a common practice to accommodate large quantities of data in deep learning, and can be used in synergy with OT.
However it amplifies the shortcomings of OT due to its marginal constraints combined with subsampling effects, which is detrimental to learning application performances. 
To mitigate this issue, we propose to relax those constraints and use UOT at the minibatch level. 
We showed that not only theoretical properties are preserved with such loss, but it also dampens negative coupling effects, yielding a more efficient measure of comparison between data distributions. We notably showed it can reach state-of-the-art performances on challenging domain adaptation problems.
We believe those results will encourage the use of minibatch Unbalanced  OT in machine learning applications.



\subsubsection*{Acknowledgements}

Authors would like to thank Youn\`es Zine, J\'er\'emy Cohen for fruitful discussions and Szymon Majewski for his proof checking. This work is partially funded through the projects OATMIL ANR-17-CE23-0012, OTTOPIA ANR-20-CHIA-0030 and 3IA Côte d'Azur Investments ANR-19-P3IA-0002 of the French National Research Agency (ANR). This research was produced within the framework of Energy4Climate Interdisciplinary Center (E4C) of IP Paris and Ecole des Ponts ParisTech. This research was supported by 3rd Programme d’Investissements d’Avenir ANR-18-EUR-0006-02. This action benefited from the support of the Chair "Challenging Technology for Responsible Energy" led by l’X – Ecole polytechnique and the Fondation de l’Ecole polytechnique, sponsored by TOTAL. 

\bibliographystyle{icml2021}
\bibliography{egbib}

\appendix
\onecolumn

{\centering{\LARGE\bfseries Unbalanced minibatch Optimal Transport; applications to Domain Adaptation}

\vspace{1em}
\centering{{\LARGE\bfseries Supplementary material}}

}

\paragraph{Outline.} The supplementary material of this paper is organized as follows:
\begin{itemize}
    \item In section A, we first review the formalism with definitions and basic property proofs.
    \item In section B, we demonstrate our statistical and optimization results.
    \item In section C, we give extra experiments and details for domain adaptation experiments.
\end{itemize}


\section{Minibatch UOT formalism and basic properties}
We start with the rigorous formalism of the minibatch UOT transport plan.

\subsection{Minibatch UOT plan formalism}
\begin{definition}
We denote by $\operatorname{Opt}_h$ the set of all optimal transport plans for $h=\operatorname{OT}_\phi^{\tau, \varepsilon}$, cost matrix $C$ and a marginal $\uu$. Let $\mathbf{u}_m, \mathbf{u}_{m} \in (\mathbb{R}^m)^2$ be discrete positive uniform vectors.
For each pair of index $m$-tuples $I = (i_1,\ldots,i_m)$ and $J = (j_1,\ldots,j_m)$ from $\llbracket 1,n\rrbracket^m$, consider $C' := C_{I, J}$ the $m \times m$ matrix with entries $C'_{k\ell} = C_{i_k,j_\ell}$ and denote by $\Pi^{m} _{ I, J }$ an arbitrary element of $\operatorname{Opt}_h$.
It can be lifted to an $n \times n$ matrix where all entries are zero except those indexed in $I \times J$:
\begin{align}
    \label{eq:minibatchTPlifted}
    \Pi_{I,J} &= Q_I^\top \Pi^{m}_{I,J} Q_J\\
\intertext{where $Q_I$ and $Q_J$ are $m \times n$ matrices defined entrywise as}
(Q_I)_{k i} &= \delta_{i_k,i}, 1 \leq k \leq m, 1 \leq i \leq n\\
(Q_J)_{\ell j} &= \delta_{j_\ell,j}, 1 \leq \ell \leq m, 1 \leq j \leq n.
\end{align}
Each row of these matrices is a Dirac vector, hence they satisfy $Q_I \mathbf{1}_n = \mathbf{1}_m$ and $Q_J \mathbf{1}_n = \mathbf{1}_m$.
\end{definition}

We also define the \textit{averaged minibatch transport matrix} which takes into account all possible minibatch couples.

\begin{definition}[Averaged minibatch transport matrix]
\label{def:AVG_OT_plan}

Consider $h=\operatorname{OT}_\phi^{\tau, \varepsilon}$. Given data $n$-tuples $\XX,\YY$, consider for each pair of $m$-tuples $I$, $J$, the uniform vector of size $m, \uu_m$, and let $\Pi_{I,J}$ be defined as in Definition~\ref{eq:minibatchTPlifted}.

The averaged minibatch transport matrix and its incomplete variant are :
\begin{align}
 &\overline{\Pi}^m(\XX, \YY)  \defas \frac{(n-m)!^2}{n!^2} \sum_{I \in \PP} \sum_{J \in \PP} \Pi_{  I, J }. \label{eq:pim},\\
  &\widetilde{\Pi}_k^m(\XX, \YY) :=  k^{-1} \sum_{  (I, J)  \in D_k  }  \Pi_{  I, J },
\end{align}
where $D_k$ is a set of cardinality $k$ whose elements are drawn at random from
the uniform distribution on $ \Gamma:= \mathcal{P}_m \times \mathcal{P}_m  $.
\end{definition}
The average in the above definition is always finite so we do not need to concern ourselves with the measurability of selection of optimal transport plans. The same will be true whenever an average of optimal transport plans will be taken in the rest of this paper, since all results concerning such averages will be nonasymptotic. We will therefore avoid further mentioning this issue, for the sake of brevity. Unfortunately, on contrary to the balanced case, the minibatch UOT transport plan do not define OT transport plan as they do not respect the marginals, so in general the averaged minibatch UOT is not an OT transport plan. Note that the Sinkhorn divergence involves three terms, which explains why we can not define an associated averaged minibatch transport matrix.

\subsection{Basic properties}
\begin{proposition}[Positivity, symmetry and bias]
  The minibatch UOT are positive and symmetric losses. However, they are not definites, \emph{i.e.,} $\bar{h}^m(\XX, \XX) > 0$ for non trivial $\XX$ and $1 < m < n$.

\begin{proof}
The first two properties are inherited from the classical UOT cost. Consider a uniform probability vector and random $3$-data tuple $\XX=(\xx_1, \xx_2, \xx_3 )$ with distinct vectors $(\xx_i)_{1\leq i\leq 3}$. As $\bar{h}^m$ is an average of positive terms, it is equal to 0 if and only if each of its term is 0. But consider the minibatch term $I_1 = ( i_1, i_2 )$ and $ I_2 = ( i_1, i_3 )$, then obviously $h(\uu, \uu, C(\XX(I_1), \XX(I_2))) \ne 0$ as $\xx_2 \ne \xx_3$, where $\XX(I_1)$ denotes the data minibatch corresponding to indices in $I_1$.
\end{proof}
\end{proposition}

We now give the proof for our claim "A simple combinatorial argument assures that the sum  of $\uu_m$ over all $m$-tuples $I$ gives $\uu_n$."
\begin{proposition}[Averaged distributions]
  Let $\uu_m$ be a uniform vector of size $m$. The average over $m$-tuples $I \in \PP$ for a given index of $\uu_m$ is equal to $\frac{m_\aa}{n}$, i.e., $\forall i \in \llbracket 1,n \rrbracket, \sum_{I \in \PP} (\uu_m)_i = (\uu_n)_i = \frac{m_\aa}{n}$.
\end{proposition}

\begin{proof}
  We recall that $\PP$ denotes the set of all $m$-tuples without repeated elements. Let us check we recover the initial weights $(\uu_n)_i = \frac{m_\aa}{n}$.
Observe that $\sum_{i=1}^n a_i=m_\aa$ and that for each $1 \leq i \leq n$
\begin{align}
\sharp\{I \in \PP: i \in I\}&=\sharp\{I \in \PP: n \in I\}\nonumber \\
& = \sharp \{ I=(i_{1},\ldots,i_{m}) \in \PP: i_{1}=n\} + \ldots + \sharp \{ I = (i_{1},\ldots,i_{m}) \in \PP: i_{m}=n\}\nonumber \\
&= m \cdot \sharp \{ I = (i_{1},\ldots,i_{m}) \in \PP: i_{m}=n\}.\label{app:num_i_all_tup}
\end{align}
Since $\sharp \{ I = (i_{1},\ldots,i_{m}) \in \PP: i_{m}=n\}$ is the number of $(m-1)$-tuples without repeated indices of $\llbracket 1,n-1\rrbracket$, $(n-1)!/(n-m)!$, it follows that
\begin{align}
 \frac{(n-m)!}{n!} \cdot \sum_{I \in \PP} \frac{m_\aa}{m} 1_I(i)
&= \frac{(n-m)!}{n!} \sum_{I \in \PP, i \in I} \frac{m_\aa}{m}
=
\frac{(n-m)!}{n!} \frac{m_\aa}{m} \cdot \sharp\{I \in \PP: i \in I\}\\
&=
\frac{(n-m)!}{n!} \frac{m_\aa}{m} m \cdot \frac{(n-1)!}{(n-m)!}
=
\frac{m_\aa}{n}
\end{align}
\end{proof}

\section{Proof main results}
In this section we prove the UOT properties and the minibatch statistical and optimization theorems. We start with UOT properties as they are necessary to derive the minibatch results. 
\subsection{Unbalanced Optimal Transport properties}

We recall the definition of Csiszàr divergences. Consider a convex, positive, lower-semicontinuous function such that $\phi(1)=0$. Define its recession constant as $\phi^\prime_\infty = \lim_{x\rightarrow +\infty} \phi(\xx) / \xx$. The Csiszàr divergence between positively weighted vectors $(\xx,\yy)\in \mathbb{R}_+^d$ reads
\begin{align*}
D_\phi(\xx, \yy) = \sum_{\yy_i\neq 0} \yy_i \phi\Big( \frac{\xx_i}{\yy_i} \Big) + \phi^\prime_\infty \sum_{\yy_i = 0} \xx_i.
\end{align*}

It allows to generalize OT programs. We retrieve common penalties such as Total Variation and Kullback-Leibler divergence by respectively taking $\phi(\xx) = |\xx-1|$ and $\phi(\xx) = (\xx\log \xx - \xx + 1)$.
We provide a generalized definition of all OT programs as

\begin{align*}
	\operatorname{OT}_\phi^{\tau, \varepsilon}(\aa, \bb, C) &= \underset{\Pi \in \mathbb{R}_+^{n \times n}}{\text{min}} \mathcal{F}(\Pi, C)= \underset{\Pi \in \mathbb{R}_+^{n \times n}}{\text{min}} \langle C, \Pi \rangle + \tau D_\phi(\Pi\mathbf{1}_n|\aa) + \tau D_\phi(\Pi^T\mathbf{1}_n|\bb) + \varepsilon \texttt{KL}(\Pi|\aa \otimes \bb).
\end{align*}

Where $\mathcal{F}$ denotes the UOT energy. 
\subsubsection{Robustness}

We start by showing the robustness properties lemma 1 that we split in two different lemmas. Lemma 1.1 shows that the UOT cost is robust to an outlier while lemma 1.2 shows that OT is not robust to an outlier.

\begin{customlemma}{1.1}
	Take $(\mu,\nu)$ two probability measures with compact support, and $z$ outside of $\nu$'s support. Recall the Gaussian-Hellinger distance~\cite{Liero_2017} between two positive measures as
	\begin{align*}
	\operatorname{GH}_\tau(\mu,\nu) = \inf_{\pi\geq 0} \int C(x,y) d\pi(x,y) + \tau \texttt{KL}(\pi_1|\mu) + \tau \texttt{KL}(\pi_2|\nu).
	\end{align*}
	For $\zeta\in[0,1]$, write $\tilde{\mu} = \zeta\mu + (1-\zeta)\delta_z$ a measure perturbed by a Dirac outlier. Write $m(z) = \int C(z,y) d\nu(y)$ One has
	\begin{align}\label{eq-bound-uot-outlier}
		\operatorname{GH}_\tau(\tilde{\mu},\nu) \leq \zeta \operatorname{GH}_\tau(\mu,\nu) + 2\tau(1-\zeta)(1 - e^{-m(z) / 2\tau})
	\end{align}
	
	In particular, with the notation $\operatorname{OT}_\texttt{KL}^{\tau, 0}$ it reads
	\begin{align}
		\operatorname{OT}_\texttt{KL}^{\tau, 0}(\tilde{\mu}, \nu, C) &\leq \zeta \operatorname{OT}_\texttt{KL}^{\tau, 0}(\mu, \nu,C) + 2\tau(1-\zeta)(1 - e^{-m(\zz) / 2\tau})\nonumber
	\end{align}
\end{customlemma}
\begin{proof}
	Recall that the OT program reads
	
	Write $\pi$ the optimal plan for $\operatorname{OT}_\phi^{\tau, 0}(\mu,\nu)$. We consider a suboptimal plan for $\operatorname{OT}_\phi^{\tau, 0}(\tilde{\mu}, \nu)$ of the form
	\begin{align*}
		\tilde{\pi} = \zeta \pi + (1-\zeta)\kappa \delta_z\otimes\nu,
	\end{align*}
	where $\kappa$ is mass parameter which will be optimized after. Note that the marginals of the plan $\tilde{\pi}$ are
	$\tilde{\pi}_1 = \zeta\pi_1 + (1-\zeta)\kappa\delta_z$ and $\tilde{\pi}_2 = \zeta\pi_2 + (1-\zeta)\kappa\nu$.
	Note that \texttt{KL} is jointly convex, thus one has
	\begin{align*}
		\texttt{KL}(\tilde{\pi}_1|\tilde{\mu})&\leq \zeta \texttt{KL}(\pi_1|\mu) + (1-\zeta)\texttt{KL}(\kappa\delta_z|\delta_z),\\
		\texttt{KL}(\tilde{\pi}_2|\tilde{\mu})&\leq \zeta \texttt{KL}(\pi_2|\nu) + (1-\zeta)\texttt{KL}(\kappa\nu|\nu).
	\end{align*}
	Thus a convex inequality yields
	\begin{align*}
		\operatorname{OT}_\phi^{\tau, 0}(\tilde{\mu}, \nu)&\leq
				\zeta \Big[\int ||x-y|| d\pi(x,y) +\tau \texttt{KL}(\pi_1|\mu) + \tau \texttt{KL}(\pi_2|\nu) \Big]\\
				&+ (1-\zeta) \Big[ \kappa m(z) + \tau \texttt{KL}(\kappa\delta_z|\delta_z) + \tau \texttt{KL}(\kappa\nu|\nu) \Big].
	\end{align*}
	We optimize now the upper bound w.r.t. $\kappa$. Both \texttt{KL} terms are equal to $\phi(\kappa)=\kappa\log\kappa - \kappa + 1$, thus differentiating w.r.t. $\kappa$ yields
	\begin{align*}
		m(z) + 2\tau\log\kappa = 0 \Rightarrow \kappa = e^{-m(z) / 2\tau}.
	\end{align*}
	Reusing this expression of $\kappa$ in the upper bound yields Equation~\eqref{eq-bound-uot-outlier}.
\end{proof}

\begin{customlemma}{1.2}
	Take $(\mu,\nu)$ two probability measures with compact support, and $z$ outside of $\nu$'s support. Define the Wasserstein distance between two probabilities as
	\begin{align}
		\operatorname{W}(\mu,\nu) = \sup_{f(x)+g(y)\leq C(x,y)} \int f(x) d\mu(x) + \int g(y)d\nu(y).
	\end{align}
	For $\zeta\in[0,1]$, write $\tilde{\mu} = \zeta\mu + (1-\zeta)\delta_z$ a measure perturbed by a Dirac outlier. Write $(f,g)$ the optimal dual potentials of $\operatorname{W}(\mu,\nu)$, and $y^*$ a point in $\nu$'s support. One has
	\begin{align}\label{eq-bound-ot-outlier}
	\operatorname{W}(\tilde{\mu},\nu) \geq \zeta \operatorname{W}(\mu,\nu)
	+ (1-\zeta)\Big(||z - y^*||^2 - g(y^*) + \int gd\nu \Big)
	\end{align}
	In particular, with the notation $\operatorname{OT}_\texttt{KL}^{\infty, 0}$ it reads
	\begin{align}
		\operatorname{OT}_\phi^{\infty, 0}(\tilde{\mu}, \nu) &\geq \zeta \operatorname{OT}_\phi^{\infty, 0}(\mu, \nu) + (1-\zeta)\Big(C(\zz, y^*) - g(y^*) + \int gd\beta\Big)\nonumber
	\end{align}
\end{customlemma}
\begin{proof}
	We consider a suboptimal pair $(\tilde{f},\tilde{g})$ of potentials for $\operatorname{OT}_\phi^{\infty, 0}(\tilde{\mu}, \nu)$. On the support of $(\mu,\nu)$ we take the optimal potentials pair $(f,g)$ for $(\mu,\nu)$, i.e. $\tilde{f}=f$ and $\tilde{g}=g$. We need to extend $\tilde{f}$ at $z$. To do so we take the $c$-transform of $g$, i.e.
	\begin{align*}
		\tilde{f}(z) = \inf_{y\in spt(\nu)} ||z - y||^2 - g(y) = ||z - y^*||^2 - g(y^*),
	\end{align*}
	where the infimum is attained at some $y^*$ since $\nu$ has compact support. the pair $(\tilde{f}, \tilde{g})$ is suboptimal, thus
	\begin{align*}
	\operatorname{OT}_\phi^{\infty, 0}(\tilde{\mu}, \nu) &\geq \int \tilde{f}(x) d\tilde{\mu}(x) + \int \tilde{g}(y) d\nu(y)\\
	&\geq \zeta\int f(x) d\mu(x) + (1-\zeta)\tilde{f}(z) + \int \tilde{g}(y)\\
	&\geq \zeta \operatorname{OT}_\phi^{\infty, 0}(\mu, \nu) + (1-\zeta)\Big[ ||z - y^*||^2 - g(y^*) + \int g(y) d\nu(y) \Big]
	\end{align*}
	Hence the resulted given by Equation~\eqref{eq-bound-ot-outlier}.
\end{proof}

\subsubsection{UOT properties}
Now let us present results which will be useful for concentration bounds. A key element is to have a bounded plan and a finite UOT cost in order to derive a hoeffding type bound. We start this section by proving lemma 2. We split it in two, lemma 2.1 proves that the UOT cost is finite and provides an upper bound while lemma 2.2 proves that the UOT plan exists and belongs to a compact set.

\begin{customlemma}{2.1}[Upper bounds]\label{app:upper_bound_UOT} Let $(\aa,\bb)$ be two positive vectors and assume that $\langle \aa\bb^\top, C \rangle < +\infty$, then the UOT cost is finite. Furthermore, we have the following bound for $h = \operatorname{OT}_\phi^{\tau, \varepsilon}$, one has $\V h( \aa, \bb, C ) \V  \leqslant M_{\aa,\bb}^h,$ where
\begin{equation}
  M_{\aa,\bb}^h =  M m_\aa m_\bb  + \tau m_\aa \phi(m_\bb) + \tau m_\bb \phi(m_\aa).
\end{equation}\label{app:upper_bound}
Regarding $h = S_\phi^{\tau, \varepsilon}$, one has $\V   h( \aa, \bb, C ) \V  \leqslant M_{\aa,\bb}^S,$ where
\begin{equation}
  M_{\aa,\bb}^S = 2M m_\aa m_\bb  + \tau m_\aa \phi(m_\bb) +\tau m_\bb \phi(m_\aa) + \tau m_\aa \phi(m_\aa) +\tau m_\bb \phi(m_\bb) +\frac{\varepsilon}{2}(m_\aa - m_\bb)^2.
\end{equation}
\end{customlemma}
\begin{proof} As $\langle \aa\bb^\top, C \rangle < +\infty$ is finite, one can bound the ground cost $C$ as $ 0  \leqslant C_{i,j}  \leqslant M$. Consider the OT kernel $h\in \{ \operatorname{OT}_\phi^{\tau, \varepsilon} \}$ for any $ \varepsilon \geq 0$. Let us consider the transport plan $\Pi = \aa\bb^\top = (a_i b_j)$ (with respect to the cost matrix $C$). Because all terms are positive, we have:
\begin{align}
\V h \V  &\leqslant \langle \aa\bb^\top, C \rangle + \varepsilon \texttt{KL}(\aa\bb^\top|\aa\bb^\top) + \tau D_\phi\left((\aa\bb^\top) \mathbf{1}_n|\aa \right)+ \tau D_\phi\left((\bb\aa^\top) \mathbf{1}_n|\bb \right) \nonumber \\
&\leqslant M \sum_{i,j}a_i b_j  + \tau m_\aa \phi(m_\bb) +\tau m_\bb \phi(m_\aa) \nonumber \\
&\leqslant  M m_\aa m_\bb  + \tau m_\aa \phi(m_\bb) +\tau m_\bb \phi(m_\aa)
\end{align}
Defining $M_{\aa,\bb}^h $ as the last upper bound finishes the proof.
The case $h = S_\phi^{\tau, \varepsilon}$, is the sum of three terms of the form $ \operatorname{OT}_\phi^{\tau, \varepsilon} $ Thus the sum $ M_{\aa,\bb}^h + \tfrac{1}{2}M_{\aa,\aa}^h+ \tfrac{1}{2}M_{\bb,\bb}^h$ is an upper bound of $S_{\varepsilon}$.
\end{proof}

We now bound the UOT plan.

\begin{customlemma}{2.2}[locally compact optimal transport plan]\label{app:lemma2_2}
	Assume that $\langle \aa\bb^\top, C \rangle < +\infty$. Consider regularized or unregularized UOT with entropy $\phi$ and penalty $D_\phi$ such that one has $\phi^\prime_\infty >0$. Then there exists an open neigbhourhood $U$ around $C$, and a compact set $K$, such that the set of optimal transport plan for any $\tilde{C} \in U$ is in K, \emph{i.e.,} $\operatorname{Opt}_h(\tilde{C}) \subset K$. Furthermore, if all costs are uniformly bounded such that $0 \leq C \leq M < \infty$, then the compact K can be taken global, i.e. independent of $C$.
\end{customlemma}
\begin{proof}
  We identify the mass of a positive measure with its L1 norm, i.e. $m_\aa = \sum \aa_i = ||\aa||_1$.
  We first consider the case where $0 \leq C \leq M < \infty$.
	The OT cost is finite because the plan $\pi = \aa\bb^\top$ is suboptimal and yields $\operatorname{OT}_\phi^{\tau, \varepsilon}(\aa, \bb, C) \leq M m_\aa m_\bb + \tau m_\aa \phi(m_\bb) + \tau m_\bb \phi(m_\aa) < +\infty$.

	Take a sequence $\Pi_t$ approaching the infimum. Note that thanks to the Jensen inequality, one has $D_\phi(\xx,\yy) \geq m_\yy \phi(m_\xx / m_\yy)$ (see~\cite{Liero_2017}). Write $m_\Pi = \sum \Pi_{t,ij}$. One has for any $t$
	\begin{align*}
	\langle \Pi_t, C \rangle &+ \tau D_\phi(\Pi_{t,1}\mathbf{1}_n|\aa) + \tau D_\phi(\Pi_{t,2}^\top \mathbf{1}_n|\bb) + \epsilon\textrm{KL}(\Pi_t| \aa\bb^\top)\\
	&\geq m_\Pi \Bigg[ \min C_{ij} + \tau \frac{m_\aa}{m_\Pi}\phi \big(\frac{m_\Pi}{m_\aa} \big)
	+ \tau \frac{m_\bb}{m_\Pi}\phi \big(\frac{m_\Pi}{m_\bb} \big)
	+ \varepsilon\frac{m_\aa m_\bb}{m_\Pi}\phi_{KL} \big(\frac{m_\Pi}{m_\aa m_\bb} \big)\Bigg]\\
	&\geq m_\Pi L(m_\Pi).
	\end{align*}
	If $||\Pi_t||_1 = m_\Pi \rightarrow +\infty$, then $L(m_\Pi)\rightarrow +\infty$ if $\varepsilon>0$ and $L(m_\Pi)\rightarrow \min C_{ij} + 2\phi^\prime_\infty>0$ otherwise.
	In both cases, as $t\rightarrow \infty$ and $||\Pi_t||_1 = m_\Pi \rightarrow +\infty$, we are supposed to approach the infimum but its lower bound goes to $+\infty$, which contradicts the fact that the optimal OT cost is finite.
	
	More precisely, there exists a large enough value $\tilde{M}$ such that for $m_\Pi > \tilde{M}$, the lower bound is superior to the upper bound $M m_\aa m_\bb + \tau m_\aa \phi(m_\bb) + \tau m_\bb \phi(m_\aa)$ and thus necessarily not optimal. Furthermore, $\tilde{M}$ depends on $(m_\aa, m_\bb, M)$ since $0\leq C \leq M$.
	Thus, there exists $\tilde{M} >0$ and some $t_0$ such that for $t\geq t_0$ any plan approaching the optimum statisfies $||\Pi_t||_1\leq \tilde{M}$. The sequence $(\Pi_t)_t$ is in a finite dimensional, bounded, and closed set, i.e. a compact set.
	One can extract a converging subsequence whose limit is a plan attaining the minimum. Thus any optimal plan is necessarily in a compact set.

	To generalize to local compactness, we consider $\delta >0$ and a neighbourhood $U$ of $C$ such that for any $\tilde{C} \in U$ one has $0 \leq \tilde{C} \leq \max C + \delta$. Reusing the above proof yields the existence of $\tilde{M}$ such that  for any $\tilde{C} \in U$, any plan approaching the optimum satisfies $||\Pi_t||_1\leq \tilde{M}$, but this time $\tilde{M}$ depends on $(m_\aa, m_\bb, \max C + \delta)$, which is independent of $C$ in its neighbourhood.
\end{proof}

\long\def\comment#1{
\begin{customlemma}{2.2}[Bounded optimal transport plan]\label{app:lemma2_2}
	Assume that $\langle \aa\bb^\top, C \rangle < +\infty$. Consider regularized or unregularized UOT with entropy $\phi$ and penalty $D_\phi$ such that one has $\phi^\prime_\infty >0$. Then the optimal transport plan exists and can be taken into a compact set.
\end{customlemma}

\begin{proof}
	We identify the mass of a positive measure with its L1 norm, i.e. $m_\aa = \sum \aa_i = ||\aa||_1$.
	The OT cost is finite because the plan $\pi = \aa\bb^\top$ is suboptimal and yields $\operatorname{OT}_\phi^{\tau, \varepsilon}(\aa, \bb) \leq \langle \aa\bb^\top, C \rangle + \tau m_\aa \phi(m_\bb) + \tau m_\bb \phi(m_\aa) < +\infty$. Take a sequence $\Pi_t$ approaching the infimum. Note that thanks to the Jensen inequality, one has $D_\phi(\xx,\yy) \geq m_\yy \phi(m_\xx / m_\yy)$ (see~\cite{Liero_2017}). Write $m_\Pi = \sum \Pi_{t,ij}$. One has for any $t$
	\begin{align*}
	\langle \Pi_t, C \rangle &+ \tau D_\phi(\Pi_{t,1}\mathbf{1}_n|\aa) + \tau D_\phi(\Pi_{t,2}^\top \mathbf{1}_n|\bb) + \epsilon\textrm{KL}(\Pi_t| \aa\bb^\top)\\
	&\geq m_\Pi \Bigg[ \min C_{ij} + \tau \frac{m_\aa}{m_\Pi}\phi \big(\frac{m_\Pi}{m_\aa} \big)
	+ \tau \frac{m_\bb}{m_\Pi}\phi \big(\frac{m_\Pi}{m_\bb} \big)
	+ \varepsilon\frac{m_\aa m_\bb}{m_\Pi}\phi_{KL} \big(\frac{m_\Pi}{m_\aa m_\bb} \big)\Bigg].
\end{align*}
	If $||\Pi_t||_1 = m_\Pi \rightarrow +\infty$, then the right-hand term converges to $+\infty$ if $\varepsilon>0$ and to $\min C_{ij} + 2\phi^\prime_\infty>0$ otherwise. In both cases, as $t\rightarrow \infty$, we are supposed to approach the infimum but its lower bound goes to $+\infty$, which contradicts the fact that the optimal OT cost is finite. Thus, there exists $M >0$ such that $||\Pi_t||_1\leq M$. The sequence $(\Pi_t)_t$ is in a finite dimensional, bounded, and closed set, i.e. a compact set. One can extract a converging subsequence whose limit is a plan attaining the minimum. Thus any optimal plan is necessarily in a compact set, which ends the proof.
\end{proof}}

We recall we denote the set of all optimal transport plan $\operatorname{Opt}_h(\XX, \YY) \subset \mathcal{M}_+(\mathcal{X})$. While the UOT energy takes positive vectors and a ground cost as inputs, we make a slight abuse of notation with $\operatorname{Opt}_h(\XX, \YY)$. Indeed, the ground cost can be deduced from $\XX, \YY$ and we associate uniform vectors as $\aa$ and $\bb$. As each element $\Pi$ of $\operatorname{Opt}_h(\XX, \YY)$ is bounded by a constant $M$, $\operatorname{Opt}_h(\XX, \YY)$ is a compact space of $\mathcal{M}_+(\mathcal{X})$. We denote the maximal constant $M$ which bounds all elements of $\operatorname{Opt}_h(\XX, \YY)$ as $\mathfrak{M}_\Pi$. We now prove that the set of optimal transport plan is convex , which will be useful for the optimization section.

\begin{customlemma}{3}[optimal transport plan convexity]\label{app:convexity_plan_set}
  Consider regularized or unregularized UOT with entropy $\phi$ and penalty $D_\phi$. The set of all optimal transport plan $\operatorname{Opt}_h(\XX, \YY)$ is a convex set.
\end{customlemma}
\begin{proof}
It is a general property of convex analysis. Take a convex function $f$ and two points $(\xx,\yy)$ that both attain the minimum over a convex set $E$. Write $\zz=t\xx +(1-t)\yy$ for $t\in[0,1]$. By convexity and suboptimality of $\zz$ one has $\min_E f \leq f(\zz) \leq tf(\xx) + (1-t)f(\yy) = \min_E f$. Thus $z$ is also optimal, hence the set of minimizers is convex. The losses $ \operatorname{OT}_\phi^{\tau, \varepsilon}$ fall under this setting.
\end{proof}

Finally, we provide a final result about UOT cost which is also useful for the optimization properties.

\begin{customlemma}{4}[UOT is Lipschitz in the cost $C$]\label{lemma:Lipschitz}
The map $C \mapsto h(\uu,\uu, C)$ is locally Lipshitz. Furthermore, if the costs are uniformly bounded ($0\leq C\leq M$) then the loss is globally Lipschitz.
\begin{proof}
We recall that $h(\uu, \uu, C) = \underset{\Pi \in \mathbb{R}_+^{n \times n}}{\text{min}} \mathcal{F}(\Pi, C)$. Let $C_1$ and $C_2$ be two ground costs. Let $\Pi_1$ and $\Pi_2$ be the optimal solutions of $h(\uu,\uu, C_1)$ and $h(\uu,\uu, C_2)$, \emph{i.e.,} $h(\uu, \uu, C_1) = \mathcal{F}(\Pi_1, C_1)$. Then we have:
\begin{align}
\mathcal{F}(\Pi_1, C_1) - \mathcal{F}(\Pi_1, C_2) \leq h(\uu,\uu, C_1) - h(\uu,\uu, C_2) \leq \mathcal{F}(\Pi_2, C_1) - \mathcal{F}(\Pi_2, C_2)
\end{align}

Thus we have 
\begin{align}
h(\uu,\uu, C_1) - h(\uu,\uu, C_2) &\leq \mathcal{F}(\Pi_2, C_1) - \mathcal{F}(\Pi_2, C_2) \nonumber \\
& = \langle  \Pi_2, C_1 - C_2 \rangle \\
& \leq \| \Pi_2 \| \|C_1 - C_2\|
\end{align}
Where the last inequality uses the Cauchy-Schwarz inequality. Following the same logic we get a bound for minus the left hand term
\begin{align}
h(\uu,\uu, C_2) - h(\uu,\uu, C_1) \leq \mathcal{F}(\Pi_1, C_2) - \mathcal{F}(\Pi_1, C_1) \leq \| \Pi_1 \| \|C_1 - C_2\|
\end{align}
It remains to bound $\|\Pi_i\|$. When we study the local Lipschitz property, without loss of generality, we fix $C_1$ and take $C_2$ in a local neighbourhood of $C_1$.
Thus Lemma~\ref{app:lemma2_2}, gives that $\|\Pi_i\|\leq \tilde{M}$, where $\tilde{M}$ only depends on $(\phi, \tau,\epsilon, \aa, \bb, \max C)$, with $\aa=\bb=\uu$, i.e. it is locally independant of $C$ in its neighbourhood, hence the local Lipschitz property. 
When $0\leq C \leq M$, then $\tilde{M}$ is independent of the cost, hence the bound is global and the map is globally Lipschitz. 
\end{proof}
\end{customlemma}

\long\def\comment#1{
\begin{customlemma}{4}[UOT is Lipschitz in the cost $C$]\label{lemma:Lipschitz}
The map $C \mapsto h(\uu,\uu, C)$ is globally Lipshitz (i.e. the Lipschitz constant does not depend on $C$).
\begin{proof}
We recall that $h(\uu, \uu, C) = \underset{\Pi \in \mathbb{R}_+^{n \times n}}{\text{min}} \mathcal{F}(\Pi, C)$. Let $C_1$ and $C_2$ be two ground costs. Let $\Pi_1$ and $\Pi_2$ be the optimal solutions of $h(\uu,\uu, C_1)$ and $h(\uu,\uu, C_2)$, \emph{i.e.,} $h(\uu, \uu, C_1) = \mathcal{F}(\Pi_1, C_1)$. Then we have:
\begin{align}
\mathcal{F}(\Pi_1, C_1) - \mathcal{F}(\Pi_1, C_2) \leq h(\uu,\uu, C_1) - h(\uu,\uu, C_2) \leq \mathcal{F}(\Pi_2, C_1) - \mathcal{F}(\Pi_2, C_2)
\end{align}

Thus we have 
\begin{align}
h(\uu,\uu, C_1) - h(\uu,\uu, C_2) &\leq \mathcal{F}(\Pi_2, C_1) - \mathcal{F}(\Pi_2, C_2) \nonumber \\
& = \langle  \Pi_2, C_1 - C_2 \rangle \\
& \leq \| \Pi_2 \| \|C_1 - C_2\|
\end{align}
Where the last inequality uses the Cauchy-Schwarz inequality. Following the same logic we get a bound for minus the left hand term
\begin{align}
h(\uu,\uu, C_2) - h(\uu,\uu, C_1) \leq \mathcal{F}(\Pi_1, C_2) - \mathcal{F}(\Pi_1, C_1) \leq \| \Pi_1 \| \|C_1 - C_2\|
\end{align}
It remains to bound $\|\Pi_i\|$. Thanks to Lemma~\ref{app:lemma2_2}, we have that all optimal plans can be taken into a compact which satisfies $\|\Pi_i\|\leq M$, where $M$ only depends on $(\phi, \tau,\epsilon, \aa, \bb)$, with $\aa=\bb=\uu$. Hence we get the global Lipschitz property w.r.t. $C$. 
\end{proof}
\end{customlemma}}

\subsection{Statistical and optimization proofs}

We consider a positive, symmetric, definite and $\mathbf{C}^1$ ground cost and without loss of generality, we consider our ground cost to be squared euclidean. We recall our definitions and hypothesis. As the distributions $ \a$ and $ \beta $ are compactly supported, there exists a constant $M >0$ such that for any $1 \leqslant i,j \leqslant n$, $ c(\xx_i, \yy_j) \leqslant M $ with $M := \text{diam}(\text{Supp}(\a) \cup \text{Supp}(\beta))^2$. We also furthermore suppose that the input masses $m_\aa$ and $m_\bb$ of positive vectors are strictly positive and finite, i.e., $0 < m_\aa < \infty$. These hypothesis assures us that the UOT cost is finite and that the UOT plan is bounded.

\subsubsection{Proof of Theorem 1} We now give the details of the proof of theorem 1. We separate theorem 1 in two sub theorem 1.1 and theorem 1.2. In the theorem 1.1, we show the deviation bound between $\widetilde{h}_k^m$ and $E_h$ and in theorem 1.2, we show the deviation bound between $\widetilde{\Pi}_k^m$ and $\overline{\Pi}^m$. For theorem 1.1, we rely on two lemmas. The first lemma bounds the deviation between the complete estimator $\bar{h}^m$ and its expectation $E_h$. We denote the floor function as $\lfloor x \rfloor$ which returns the biggest integer smaller than $x$.

\begin{customlemma}{5}[U-statistics concentration bound]\label{thm:U_to_mean}
Let $ \delta \in (0,1) $, three integers $k \geq 1$ and $m \leq n$ be fixed, and two compactly supported distributions $\alpha, \beta$. Consider two $n$-tuples $\XX \sim \alpha^{\otimes n}$ and $\YY \sim \beta^{\otimes n}$ and a kernel $h\in \{ \operatorname{OT}_\phi^{\tau, \varepsilon}, S_\phi^{\tau, \varepsilon} \}$. We have a concentration bound between $\bar{h}^m(\XX, \YY)$ and the expectation over minibatches $E_h$ depending on the number of empirical data $n$
\begin{equation}
\vert \bar{h}^m(\XX, \YY) - E_h \vert \leq M_{\uu,\uu}^h \sqrt{\frac{ \log(2/\delta)}{2\lfloor n/m \rfloor}}
\end{equation}
with probability at least $1-\delta$ and where $M_{\uu,\uu}^h$ is an upper bound defined in lemma \ref{app:upper_bound_UOT}.
\end{customlemma}

\begin{proof}
$\bar{h}^m(\XX, \YY)$ is a two-sample U-statistic of order $2m$ and $E_h$ is its expectation as $\XX$ and $\YY$ are \emph{iid} random variables. $\bar{h}^m(\XX, \YY)$ is a sum of dependant variables and it is possible to rewrite $\bar{h}^m(\XX, \YY)$ as a sum of independent random variables. As $\alpha, \beta$ are compactly supported by hypothesis, the UOT loss is bounded thanks to lemma \ref{app:upper_bound}. Thus, we can apply the famous Hoeffding lemma to our U-statistic and get the desired bound. The proof can be found in \cite{Hoeffding1963} (the two sample U-statistic case is discussed in section 5.b) .
\end{proof}

The second lemma bounds the deviation between the incomplete estimator $\widetilde{h}_k^m$ and the complete estimator $\bar{h}^m$.

\begin{customlemma}{6}[Deviation bound]\label{app:inc_U_to_U} Let $ \delta \in (0,1) $, three integers $k \geq 1$ and $m \leq n$ be fixed, and two compactly supported distributions $\alpha, \beta$. Consider two $n$-tuples $\XX \sim \alpha^{\otimes n}$ and $\YY \sim \beta^{\otimes n}$ and a kernel $h\in \{ \operatorname{OT}_\phi^{\tau, \varepsilon}, S_\phi^{\tau, \varepsilon} \}$. We have a deviation bound between $\widetilde{h}_k^m(\XX, \YY)$ and $\bar{h}^m(\XX, \YY)$ depending on the number of batches $k$

\begin{equation}
\vert  \widetilde{h}_k^m(\XX, \YY) - \bar{h}^m(\XX, \YY) \vert \leqslant M_{\uu,\uu}^h \sqrt{\frac{2 \log(2/\delta)}{k}}
\end{equation}
with probability at least $1 - \delta$ and where $M_{\uu,\uu}^h$ is an upper bound defined in lemma \ref{app:upper_bound_UOT}.
\end{customlemma}
\begin{proof}
First note that $\widetilde{h}_k^m(\XX, \YY)$ is an subsample quantity of $\bar{h}^m(\XX, \YY)$. Let us consider the sequence of random variables $ ((\mathfrak{b}_l(I, J)_{ (I, J) \in \PP})_{1 \leqslant l \leqslant k} $ such that $\mathfrak{b}_l(I, J)  $ is equal to $1$ if $(I, J) $ has been selected at the $l-$th draw and $0$ otherwise. By construction of $ \widetilde{h}_k^m $, the aforementioned sequence is an i.i.d sequence of random vectors and the $ \mathfrak{b}_l(I, J) $ are Bernoulli random variables of parameter $ 1/ \vert \Gamma \vert $. We then have
\begin{equation}
\widetilde{h}_k^m(\XX, \YY) - \bar{h}^m(\XX, \YY) = \frac{1}{k} \sum_{l=1}^k \omega_l
\end{equation}
where $ \omega_l = \sum_{(I, J) \in \PP} ( \mathfrak{b}_l(I, J) - \frac{1}{ \vert \Gamma \vert }  ) {  h(I,J) } $. Conditioned upon $ \XX = (\xx_1, \cdots, \xx_n)$ and $\YY = (\yy_1, \cdots, \yy_n) $, the variables $ \omega_l $ are independent, centered { and bounded by $2M_{\uu,\uu}^h$ thanks to lemma \ref{app:upper_bound_UOT}}. Using Hoeffding's inequality yields
\begin{align}
\mathbb{P}( \vert  \widetilde{h}_k^m(\XX, \YY) - \bar{h}^m(\XX, \YY) \vert > \varepsilon  ) & = \mathbb{E} [ \mathbb{P}( \vert  \widetilde{h}_k^m(\XX, \YY) - \bar{h}^m(\XX, \YY) \vert > \varepsilon \vert X,Y )] \\
& = \mathbb{E} [ \mathbb{P}( \vert \frac{1}{k} \sum_{l=1}^k \omega_l ) \vert > \varepsilon \vert X,Y )]\\
& \leqslant  \mathbb{E} [ 2 e^{\frac{-k \varepsilon^2}{2(M_{\uu,\uu}^h)^2}} ] = 2 e^{\frac{-k \varepsilon^2}{2(M_{\uu,\uu}^h)^2}}
\end{align}
which concludes the proof.
\end{proof}

We are now ready to prove Theorem \ref{app:inc_U_to_mean}. 

\begin{customtheorem}{1.1}[Maximal deviation bound]\label{app:inc_U_to_mean} Let $ \delta \in (0,1) $, three integers $k \geq 1$ and $m \leq n$ be fixed and two compactly supported distributions $\alpha, \beta$. Consider two $n$-tuples $\XX \sim \alpha^{\otimes n}$ and $\YY \sim \beta^{\otimes n}$ and a kernel $h\in \{ \operatorname{OT}_\phi^{\tau, \varepsilon}, S_\phi^{\tau, \varepsilon} \}$. We have a maximal deviation bound between $\widetilde{h}_k^m(\XX, \YY)$ and the expectation over minibatches $E_h$ depending on the number of empirical data $n$ and the number of batches $k$
  \begin{equation}
    \vert \widetilde{h}_k^m(\XX, \YY) - E_h \vert \leq M_{\uu,\uu}^h \sqrt{\frac{ \log(2/\delta)}{2\lfloor n/m \rfloor}} + M_{\uu,\uu}^h \sqrt{\frac{2 \log(2/\delta)}{k}}
  \end{equation}
with probability at least 1 - $\delta$ and where $M_{\uu,\uu}^h$ is an upper bound defined in lemma \ref{app:upper_bound_UOT}.
\end{customtheorem}
\begin{proof} Thanks to lemma \ref{app:inc_U_to_U} and \ref{thm:U_to_mean} we get
\begin{align}
\vert \widetilde{h}_k^m(\XX, \YY) - E_h \vert & \leq \vert \widetilde{h}_k^m(\XX, \YY) - \bar{h}^m(\XX, \YY) \vert + \vert \bar{h}^m(\XX, \YY) - E_h \vert \\
& \leq M_{\uu,\uu}^h \sqrt{\frac{ \log(2/\delta)}{2\lfloor n/m \rfloor}} + M_{\uu,\uu}^h \sqrt{\frac{2 \log(2/\delta)}{k}}
\end{align}
with probability at least $ 1 - (\frac{\delta}{2} + \frac{\delta}{2}) = 1 - \delta $.
\end{proof}

We now give the details of the proof of theorem \ref{app:thm1_2}. In what follows, we denote by $\Pi_{(i)}$ the $i$-th row of matrix $\Pi$. Let us denote by $ \mathbf{1} \in \R^n $ the vector whose entries are all equal to $1$.

\begin{customtheorem}{1.2}[Distance to marginals]\label{app:thm1_2}
 Let $ \delta \in (0,1) $, two integers $m \leq n$ be fixed. Consider two $n$-tuples $\XX \sim \alpha^{\otimes n}$ and $\YY \sim \beta^{\otimes n}$ and the kernel $h = \operatorname{OT}_\phi^{\tau, \varepsilon}$. For all integer $ k \geq 1 $, all $ 1 \leqslant i \leqslant n  $, with probability at least $1-\delta$ on the draw of $\XX, \YY$ and $D_k$ we have
\begin{equation}
\vert  \widetilde{\Pi}_k^m(\XX, \YY)_{(i)} \mathbf{1} - \overline{\Pi}^m(\XX, \YY)_{(i)} \mathbf{1} \vert \leqslant \mathfrak{M}_\Pi^\infty \sqrt{\frac{2 \log(2/\delta)}{k}},
\end{equation}
where $\mathfrak{M}_\Pi^\infty$ denotes an upper bound of all minibatch UOT plan.
\end{customtheorem}
\begin{proof}
Let us consider the sequence of random variables $ ((\mathfrak{b}_p(I, J)_{ (I, J) \in \Gamma})_{1 \leqslant p \leqslant k} $ such that $\mathfrak{b}_p(I, J)  $ is equal to $1$ if $(I, J) $ has been selected at the $p-$th draw and $0$ otherwise. By construction of $ \widetilde{\Pi}_k^m(\XX, \YY) $, the aforementioned sequence is an i.i.d sequence of random vectors and the $ \mathfrak{b}_p(I, J) $ are bernoulli random variables of parameter $ 1/ \vert \Gamma \vert $. We then have
\begin{equation}
\widetilde{\Pi}_k^m(\XX, \YY)_{(i)}  \mathbf{1}= \frac{1}{k} \sum_{p=1}^k \omega_p
\end{equation}
where $ \omega_p = \sum_{(I, J) \in \Gamma} \sum_{j=1}^n (\Pi_{I,J})_{i,j} \mathfrak{b}_p(I, J) $. Conditioned upon $ \XX = (\xx_1, \cdots, \xx_n)$ and $\YY = (\yy_1, \cdots, \yy_n) $, the random vectors $ \omega_p $ are independent, and thanks to lemma \ref{app:lemma2_2}, they are bounded by a constant $ \frak{M}_{\Pi} $ which is the maximum mass of all optimal minibatch unbalanced plan in $\operatorname{Opt}_h(\XX(I), \YY(J))$. We denote the maximum upper bound $\mathfrak{M}_\Pi$ of all minibatch UOT plan as $\mathfrak{M}_\Pi^\infty$.
Moreover, one can observe that $ \mathbb{E}[ \widetilde{\Pi}_k^m(\XX, \YY)_{(i)}  \mathbf{1} ] = \overline{\Pi}^m(\XX, \YY)_{(i)} \mathbf{1} $. Using Hoeffding's inequality yields
\begin{align}
\mathbb{P}( \V  \widetilde{\Pi}_k^m(\XX, \YY)_{(i)} \mathbf{1}  - \overline{\Pi}^m(\XX, \YY)_{(i)} \mathbf{1} )  \V > \varepsilon  ) & = \mathbb{E} [ \mathbb{P}(  \V \frac{1}{k} \sum_{p=1}^k \omega_p  - \mathbb{E}[ \frac{1}{k} \sum_{p=1}^k \omega_p]) \V  > \varepsilon  \vert X,Y )] \\
& \leqslant  2 e^{-2\frac{k \varepsilon^2}{(\mathfrak{M}_\Pi^\infty)^2}}
\end{align}
which concludes the proof.
\end{proof}

Note that the unbalanced Sinkhorn divergence $ S_\phi^{\tau, \varepsilon} $ involves three terms of the form $\operatorname{OT}_\phi^{\tau, \varepsilon}$, hence three transport plans, which explains why we do not attempt to define an associated averaged minibatch transport matrix.


\subsubsection{Proof of Theorem 2}
To prove the exchange of gradients and expectations over minibatches we rely on Clarke differential. We need to use this non smooth analysis tool as unregularized UOT is not differentiable. It is not differentiable because the set of optimal solutions might not be a singleton. Clarke differential are generalized gradients for locally Lipschitz function and non necessarily convex. A similar strategy was developped in \cite{fatras2021minibatch}. The key element of this section is to rewrite the original UOT problem $\operatorname{OT}_\phi^{\tau, \varepsilon}$ as:
\begin{align}
  \operatorname{OT}_\phi^{\tau, \varepsilon}(\aa, \bb, C) &= \underset{\Pi \in \mathbb{R}_+^{n \times n}}{\text{min}} \langle C, \Pi \rangle + \varepsilon \texttt{KL}(\Pi|\aa \otimes \bb) + \tau D_\phi(\Pi\mathbf{1}_n|\aa)  + \tau D_\phi(\Pi^\top\mathbf{1}_n|\bb)\\
  &= \underset{\Pi \in \operatorname{Opt}_{\operatorname{OT}_\phi^{\tau, \varepsilon}}}{\text{min}} \langle C, \Pi \rangle + \varepsilon \texttt{KL}(\Pi|\aa \otimes \bb) + \tau D_\phi(\Pi\mathbf{1}_n|\aa)  + \tau D_\phi(\Pi^\top\mathbf{1}_n|\bb),
\end{align}
Where $\operatorname{Opt}_{\operatorname{OT}_\phi^{\tau, \varepsilon}}(\XX,\YY)$ is a compact set of the set of measures $\mathcal{M}_+(\mathcal{X})$. The compact set is a key element for using Danskin like theorem (Proposition B.25 \cite{bertsekas1997nonlinear}).

 We start by recalling a basic proposition for Clarke regular function:
\begin{proposition}
  A $\mathbf{C}^1$ or convex map is Clarke regular.
\begin{proof}
see Proposition 2.3.6 \citep{clarke1990optimization}
\end{proof}
\end{proposition}

We first give a lemma which gives the Clarke regularity of the UOT cost with respect to a parametrized random vector.

\long\def\comment#1{
\begin{customlemma}{7}\label{app:OT_clark_reg}
  Let $\uu$ be uniform probability vectors. Let $\XX$ be a $\R^{dm}$-valued random variable, and $\{\YY_{\theta} \}$ a family of $\R^{dm}$-valued random variables defined on the same probability space, indexed by $\theta \in \Theta$, where $\Theta \subset \R^{q}$ is open. 
  Assume  the cost and the map $\theta \mapsto \YY_{\theta}$ to be $\mathbf{C}^1$, such that $\theta\mapsto C(X, Y_\theta)$ is also $\mathbf{C}^1$. Consider $h \in \{ \operatorname{OT}_\phi^{\tau, \varepsilon}, S_\phi^{\tau, \varepsilon}\}$. 
  Then the function $\theta \mapsto -h(\uu,\uu,C(\XX, \YY_{\theta}))$ is Clarke regular. Furthermore, for $h = \operatorname{OT}_\phi^{\tau, \varepsilon}$ and for all $1 \leq i \leq q$ we have:
\begin{align} \label{eq:exchange_theorem_eq1_proof}
\partial_{\theta_i} h(\uu,\uu,C(\XX, \YY_{\theta})) = \overline{\text{co}}\{ -\text{tr}(\Pi \cdot D^{T})\cdot (\nabla_{\theta_i} Y): & \Pi \in \operatorname{Opt}_h(C(\XX, \YY_\theta)), \\
D \in \R^{m, m}, \hspace{2pt} & D_{j,k} = \nabla_{Y} C_{j,k}(\XX, \YY_{\theta}) \} \nonumber
\end{align}
where $\partial_{\theta_i}$ is the Clare subdifferential with respect to $\theta_i$, $\nabla_Y C_{j,k}$ is the subdifferential of the cell $C_{j,k}$ of the  cost matrix with respect to $Y$, $\operatorname{Opt}_h(\XX, \YY)$ is the set of optimal transport plan and $\overline{\text{co}}$ denotes the closed convex hull.

\end{customlemma}
\begin{proof}
  We start with the Clarke regularity of $C\mapsto\operatorname{OT}_\phi^{\tau, \varepsilon}(\uu,\uu, C)$ by verifying the assumptions of Danskin's theorem~\citep[Theorem 2.1]{clarke1975generalized} since $h=\operatorname{OT}_\phi^{\tau, \varepsilon}$ is defined as a minimization program, considered here to depend on $(\Pi,C)$. 
  The energy associated to $h$ (see Equation~2) is l.s.c. in $(\Pi , C)$ thanks to~\citep[Lemma 3.9]{Liero_2017} and it is convex  in $C$ as it is linear. 
  We satisfy the conditions of~\citep[Theorem 2.1]{clarke1975generalized} which allows to apply this result and yields the Clarke regularity of $h$.
  
  We now focus on the regularity of the composition of $h$ with the $\mathbf{C}^1$ map $\theta\mapsto C(\XX, \YY_{\theta})$ and verify the assumptions of~\citep[Theorem 2.3.10]{clarke1990optimization}.
  The function $\theta\mapsto C(\XX, \YY_{\theta})$ is Clarke regular because it is $\mathbf{C}^1$. Furthermore, its derivative reads
  \[
  \nabla_{\theta_i} C_{j,k}(\XX, \YY_{\theta}) =  \nabla_Y C_{j,k}(\XX, \YY_{\theta}) \cdot \nabla_{\theta_i} Y_{\theta}.
  \]
  We proved above that $h$ is Clarke differentiable. More precisely, $h$ is a minimization of an energy which is linear in $C$, and it is thus concave in $C$, hence $-h$ is Clarke regular, we can apply the Chain rule~\citep[Theorem 2.3.10]{clarke1990optimization} (where equality of differentials in this result holds) to $\theta\mapsto\operatorname{OT}_\phi^{\tau, \varepsilon}(\uu,\uu, C(\XX, \YY_{\theta}))$. Combining the formulas of the Danskin theorem with the Chain rule yields Equation~\eqref{eq:exchange_theorem_eq1_proof}. When $\epsilon >0$ the set $\operatorname{Opt}_h(C(\XX, \YY)_\theta)$ is reduced to a singleton, and the notation $\overline{\text{co}}$ is superfluous.
  \end{proof}}

\begin{customlemma}{7}\label{app:OT_clark_reg}
  Let $\uu$ be a uniform probability vector. Let $\XX$ be a $\R^{dm}$-valued random variable, and $\{\YY_{\theta} \}$ a family of $\R^{dm}$-valued random variables defined on the same probability space, indexed by $\theta \in \Theta$, where $\Theta \subset \R^{q}$ is open. Assume that $\theta \mapsto \YY_{\theta}$ is $\mathbf{C}^1$. Consider a $\mathbf{C}^1$ cost $C$ and let $h \in \{ \operatorname{OT}_\phi^{\tau, \varepsilon}, S_\phi^{\tau, \varepsilon}\}$. Then the function $\theta \mapsto -h(\uu,\uu,C(\XX, \YY_{\theta}))$ is Clarke regular. Furthermore, for $h = \operatorname{OT}_\phi^{\tau, \varepsilon}$ and for all $1 \leq i \leq q$ we have:
\begin{align} \label{eq:exchange_theorem_eq1_proof}
\partial_{\theta_i} h(\uu,\uu,C(\XX, \YY_{\theta})) = \overline{\text{co}}\{ -\langle\Pi \cdot D\rangle \cdot (\nabla_{\theta_i} Y): & \Pi \in \operatorname{Opt}_h(\XX, \YY), \\
D \in \R^{m, m}, \hspace{2pt} & D_{j,k} = \nabla_{Y} C_{j,k}(\XX, \YY_{\theta}) \} \nonumber
\end{align}
where $\partial_{\theta_i}$ is the Clarke subdifferential with respect to $\theta_i$, $\nabla_Y C_{j,k}$ is the differential of the cell $C_{j,k}$ of the  cost matrix with respect to $Y$, $\operatorname{Opt}_h(\XX, \YY_\theta)$ is the set of optimal transport plan and $\overline{\text{co}}$ denotes the closed convex hull. Note that when $\varepsilon >0$ the set $\operatorname{Opt}_h(\XX, \YY_\theta)$ is reduced to a singleton, and the notation $\overline{\text{co}}$ is superfluous.
\end{customlemma}
\begin{proof}
  We start with the regularity of $\theta \mapsto -\operatorname{OT}_\phi^{\tau, \varepsilon}(\uu,\uu, C(\XX, \YY_{\theta}))$. To prove the Clarke regularity of this map, we rely on a chain rule argument. Consider the function $Y \mapsto C_{j,k}(\XX, \YY)$, it is Clarke regular because it is $\mathbf{C}^1$. Since $\theta \mapsto \YY_{\theta}$ is $\mathbf{C}^1$, it follows by the chain rule that $\theta \mapsto C_{j,k}(\XX, \YY_{\theta})$ is $\mathbf{C}^1$ and thus Clarke regular.
  The Unbalanced OT cost $\operatorname{OT}_\phi^{\tau, \varepsilon}$ is a minimization of an energy which is linear in $C$, and it is thus concave in $C$, hence $-\operatorname{OT}_\phi^{\tau, \varepsilon}$ is Clarke regular by convexity. Therefore from Theorem 2.3.9(i) and Proposition 2.3.1 (for $s=1$) in \citep{clarke1990optimization} it follows that $\theta \mapsto -\operatorname{OT}_\phi^{\tau, \varepsilon}(\uu,\uu, C(\XX, \YY_{\theta}))$ is Clarke regular. 

We now furnish the gradients associated to $\theta \mapsto -\operatorname{OT}_\phi^{\tau, \varepsilon}(\uu,\uu, C(\XX, \YY_{\theta}))$. By chain rule, the gradient of $\theta \mapsto C_{j,k}(\XX, \YY_{\theta})$ reads 
  \[
  \nabla_{\theta_i} C_{j,k}(\XX, \YY_{\theta}) =  \nabla_Y C_{j,k}(\XX, \YY_{\theta}) \cdot \nabla_{\theta_i} Y_{\theta}.
  \]
We now deal with the gradient of the map $C\mapsto\operatorname{OT}_\phi^{\tau, \varepsilon}(\uu,\uu, C)$ by verifying the assumptions of Danskin's theorem~\citep[Theorem 2.1]{clarke1975generalized}. We use in particular the remark below~\citep[Theorem 2.1]{clarke1975generalized} which states that the hypothesis on the map are verified if the map is \emph{u.s.c} in both variables $(\Pi, C)$ and convex in $C$. We recall that $\operatorname{Opt}_h(\XX, \YY)$ is a compact and a convex set, thanks to lemma \ref{app:convexity_plan_set} and lemma \ref{lemma:Lipschitz}. Furthermore, the energy associated to $h=\operatorname{OT}_\phi^{\tau, \varepsilon}$ is concave in the cost $C$ and \emph{l.s.c} in $(\Pi, C)$~\citep[Lemma 3.9]{Liero_2017}. From~\citep[Theorem 2.1]{clarke1975generalized} it follows that the subderivatives of the convex function $C \mapsto -h(\uu,\uu, C)$ are equal to $\operatorname{Opt}_h(\XX, \YY)$, due to the energy's linearity in $C$. Thus combining the formulas of the Danskin theorem with the Chain rule yields Equation~\eqref{eq:exchange_theorem_eq1_proof}. When $\varepsilon >0$ the set $\operatorname{Opt}_h(\XX, \YY_\theta)$ is reduced to a singleton, and the notation $\overline{\text{co}}$ is superfluous.

%

We now give the proof for the regularity of the map $\theta \mapsto S_\phi^{\tau, \varepsilon}(\uu,\uu, C(\XX, \YY_{\theta}))$ with $\varepsilon > 0$ as when $\varepsilon = 0$, we get the unregularized UOT treated in the above paragraph. We recall that $S_\phi^{\tau, \varepsilon}$ is the summation of three terms of the form $\operatorname{OT}_\phi^{\tau, \varepsilon}$. For $\varepsilon>0$ and each term of the sum, the set of optimal plans $\operatorname{Opt}_{\operatorname{OT}_\phi^{\tau, \varepsilon}}(\XX, \YY)$ is reduced to a unique element and the differential~\eqref{eq:exchange_theorem_eq1_proof} is also a singleton, thus $\operatorname{OT}_\phi^{\tau, \varepsilon}$ is differentiable. Then $S_\phi^{\tau, \varepsilon}$ is differentiable as a difference of differentiable functions. Furthermore $S_\phi^{\tau, \varepsilon}$ is also Clarke regular as a difference of differentiable functions.
\end{proof}

We finally prove theorem 2.

\begin{customtheorem}{2}
Let $\uu$ be uniform probability vectors and let $\XX, \YY, C$ be as in lemma \ref{app:OT_clark_reg}, $h \in \{ \operatorname{OT}_\phi^{\tau, \varepsilon}, S_\phi^{\tau, \varepsilon} \}$, and assume in addition that the random variables $\XX, \{Y_{\theta}\}_{\theta \in \Theta}$ are compactly supported. If for all $\theta \in \Theta$ there exists an open neighbourhood $U$, $\theta \in U \subset \Theta$, and a random variable $K_U : \Omega \rightarrow \R$ with finite expected value, such that
\begin{equation}\label{eq:integrable_condition_opt}
\Vert C(\XX(\omega), \YY_{\theta_1}(\omega)) - C(\XX(\omega), \YY_{\theta_2}(\omega) ) \Vert \leq K_U(\omega) \Vert \theta_1 - \theta_2 \Vert
\end{equation}
then we have
\begin{align} \label{eq:exchange_theorem_eq2_proof}
\partial_{\mathbf{\theta}} \expect \left[ h(\uu, \uu, C(\XX, \YY_{\theta})) \right] =  \expect \left[ \partial_{\mathbf{\theta}} h(\uu, \uu, C(\XX, \YY_{\theta})) \right].
\end{align}
with both expectation being finite. Furthermore the function $\theta \mapsto - \expect \left[ h(\uu, \uu, C(\XX, \YY_{\theta})) \right]$ is also Clarke regular.
\end{customtheorem}

\begin{proof}
  Suppose that $U \subset \Theta$ is open and $K_U$ is a function for which \eqref{eq:integrable_condition_opt} is satisfied. As data lie in compacts the ground cost $C$, which is $\mathbf{C}^1$, is in a compact $K_C$ and as the map $C \mapsto h(\uu, \uu, C)$ is locally Lipshitz by lemma \ref{lemma:Lipschitz}, there exists a uniform constant which makes the map $C \mapsto h(\uu, \uu, C)$ globally Lipshitz on the compact $K_C$. Thus, a similar bound to~\eqref{eq:integrable_condition_opt} is also satisfied for the function $h(\uu, \uu, C(\XX(\omega), \YY_{\theta}(\omega)))$. Thanks to lemma \ref{app:OT_clark_reg}, $-h(\uu,\uu,C(\XX, \YY_{\theta}))$ is Clarke regular, the interchange \eqref{eq:exchange_theorem_eq2_proof} and regularity of $\theta \mapsto -\expect[h(\uu,\uu, C(\XX, \YY_{\theta}))]$ will follow from Theorem 2.7.2 and Remark 2.3.5 \citep{clarke1990optimization}, once we establish that the expectation on the left hand side is finite. This is direct as we suppose we have compactly supported distributions and $C$ is a $\mathbf{C}^1$ cost. Indeed consider the function which is equal to $M_{\uu,\uu}^h$ on the distributions's support and which is set to 0 everywhere else. Taking the expectation on this function is finite as $M_{\uu,\uu}^h$ is finite.
\end{proof}

\long\def\comment#1{
\begin{proof}
  Suppose that $U \subset \Theta$ is open and $K_U$ is a function for which \eqref{eq:integrable_condition_opt} is satisfied. Then a similar bound is also satisfied for the function $h(\uu, \uu, C(\XX(\omega), \YY_{\theta}(\omega)))$, since the map $C \mapsto h(\uu, \uu, C)$ is Lipshitz by lemma \ref{lemma:Lipschitz}. Thanks to lemma \ref{app:OT_clark_reg}, $-h(\uu,\uu,C(\XX, \YY_{\theta}))$ is Clarke regular, the interchange \eqref{eq:exchange_theorem_eq2_proof} and regularity of $\theta \mapsto -\expect[h(\uu,\uu, C(\XX, \YY_{\theta}))]$ will follow from Theorem 2.7.2 and Remark 2.3.5 \citep{clarke1990optimization}, once we establish that the expectation on the left hand side is finite. This is direct as we suppose we have compactly supported distributions and $C$ is a $\mathbf{C}^1$ cost. Indeed consider the function which is equal to $M_{\uu,\uu}^h$ on the distributions's support and which is set to 0 everywhere else. Taking the expectation on this function is finite as $M_{\uu,\uu}^h$ is finite.
\end{proof}}

\long\def\comment#1{
This follows trivially from the standard bound:
\begin{equation} \label{eq:p_moment_bound}
\Vert \xx - \yy \Vert^p \leq 2^{p-1} (\Vert \xx \Vert^p + \Vert \yy \Vert^p)
\end{equation}
and the assumption that $\XX, \YY_{\theta}$ have finite $p$-moments.

}

\long\def\comment#1{
\subsubsection{Proof of Theorem 3 (aistats 2020)}

The main goal of this section is to give a justification of optimization for our minibatch OT losses by giving the \textbf{proof of theorem 3}. More precisely, we show that for the losses $ W_{\varepsilon} $ and $ S_{\varepsilon} $, one can exchange the gradient symbol $ \nabla $ and the expectation $  \mathbb{E} $. It shows for example that a stochastic gradient descent procedure is unbiased and as such legitimate.\\

\textbf{Main hypothesis.} We assume that the map $ \lambda \mapsto C(A, Y_{\lambda}) $ is differentiable. For instance for GANs, it is verified when the neural network in the generator is differentiable -which is the case if the nonlinear activation functions are all differentiable- and when the cost chosen in the Wasserstein distance is also differentiable. \\
We introduce the map
\[ g: (\Pi,C) \mapsto \langle \Pi, C \rangle - \varepsilon H(\Pi)
\]\\
To prove this theorem, we first define a map we will use the "Differentiation Lemma".
\begin{lemma}[Differentiation lemma] Let V be a nontrivial open set in $\R^p$ and let $\mathcal{P}$ be a probability distribution on $\R^d$. Define a map $C : \R^d\times \R^d \times V \rightarrow \R$ with the following properties:
\begin{itemize}
    \item For any $\lambda \in V, \expect_{\mathcal{P}}[\vert C(X, Y, \lambda)\vert ] < \infty$
    \item For $P$-almost all $(X, Y) \in \R^d \times \R^d$, the map $V \rightarrow \R$, $\lambda \rightarrow C(X, Y, \lambda)$ is differentiable.
    \item There exists a $P$-integrable function $\varphi : \R^d \times \R^d \rightarrow \R$ such that $\vert \partial_{\lambda} C(X, Y, \lambda) \vert \leq g(x)$ for all $\lambda \in V$.
\end{itemize}

Then, for any $\lambda \in V$ , $E_{\mathcal{P}}[\vert \partial_{\lambda} C(X, Y, \lambda) \vert ] < \infty$ and the function $\lambda \rightarrow E_{\mathcal{P}}[C(X, Y, \lambda)]$ is differentiable with differential:
\begin{equation}
  E_{\mathcal{P}} \partial_{\lambda} [C(X, Y, \lambda)] = \partial_{\lambda} E_{\mathcal{P}} [C(X, Y, \lambda)]
\end{equation}
\end{lemma}

The following result will also be useful.
\begin{lemma}[Danskin, Rockafellar] Let $g: (z,w) \in \R^d \times \R^d \to \R $ be a function. We define $\varphi: z \mapsto \max _{w \in W} g(z,w) $ where $W \subset \R^d$ is compact. We assume that for each $w \in W$, the function $ g(\cdot, w)  $ is differentiable and that $ \nabla_z g $ depends continuously on $ (z,w)$. If in addition, $g(z,w)$ is convex in $z$, and if $ \overline{z} $ is a point such that $ \operatorname{argmax}_{w \in W}g(\overline{z},w) = \{ \overline{w} \} $, then $\varphi$ is differentiable at $\overline{z} $ and verifies
\begin{equation}
\nabla \varphi (\overline{z}) = \nabla _z g(\overline{z}, \overline{w})
\end{equation}
\label{thm:Danskin1}
\end{lemma}

The last theorem shows that the entropic loss is differentiable with respect to the cost matrix. Indeed, the theorem directly applies since the problem is strongly convex. This remark enables us to obtain the following result.

\begin{theorem}[Exchange gradient and expectation] Let us suppose that we have two distributions $\alpha$ and $\beta$ on two bounded subsets $\mathcal{X}$ and $\mathcal{Y}$, a $\mathcal{C}^1$ cost. Assume $\lambda \mapsto Y_{\lambda}$ is differentiable. Then for the entropic loss and the Sinkhorn divergence:
\begin{equation}
\nabla_{\lambda} \expect_{Y_{\lambda} \sim \beta_{\lambda}^{\otimes m}}  h(A,Y_{\lambda}) =  \expect_{Y_{\lambda} \sim \beta_{\lambda}^{\otimes m}} \nabla_{\lambda} h(A,Y_{\lambda})
\label{app:exchange_grad_exp}
\end{equation}
\end{theorem}

\begin{proof}
Regarding the Sinkhorn divergence, as it is the sum of three terms of the form $ W_{\varepsilon} $, it suffices to show the theorem for $ h = W_{\varepsilon} $

The first and the third conditions of the Differentiation Lemma are trivial as we have supposed that our distributions have compact supports. Hence, the minibatch Wasserstein exists and is bounded on a finite set. We can also build a measurable function $\varphi$ which takes the biggest cost value $||c||_{\infty}$ inside $\mathcal{X}$ and 0 outside. As $\mathcal{X}$ is compact, the integral of the function over $\R^d$ is finite.\\
The problem is in the second hypothesis where we need to prove that $W_{\varepsilon}$ is differentiable almost everywhere. We have to show that the following function $ \lambda \mapsto \varphi_A(\lambda) $ is differentiable:
$$
\varphi_A : \lambda \mapsto \min_{\Pi \in U(a,b)} \langle \Pi, C(A,\lambda) \rangle - \varepsilon H(\Pi)
$$
where $ C(A,\lambda)  $ is the cost computed using pairwise distances between $A$ and $ Y_{\lambda} $. Since $ \lambda \mapsto  C(A,\lambda)  $ is differentiable almost everywhere by our hypothesis on $ \lambda \mapsto y_{\lambda} $, it suffices, by composition, to show that $ C \mapsto  \min_{\Pi \in U(a,b)} \langle \Pi, C \rangle - \varepsilon H(\Pi) $ is differentiable in $ C \in \R^{m \times m} $. We obtain this using lemma \ref{thm:Danskin1}. We can apply lemma \ref{thm:Danskin1} because theorem 2 gives us that the optimal transport plan belongs to a compact set of all joint measures and the fact that there is one unique solution to the entropically regularized optimal transport problem.

\end{proof}
}

\section{Domain adaptation and partial domain adaptation experiments}

In this section we provide architecture and training procedure details for the domain adaptation experiments. We also discuss the reported scores procedure. Then, we provide a parameter sensitivity analysis on our method. Finally we discuss the training behaviour for both \textsc{jumbot} and \textsc{deepjdot}. 

\subsection{Domain adaptation}

\long\def\comment#1{
  \textbf{VisDA-2017} \cite{visda} is a large-scale dataset and
  challenge for unsupervised domain adaptation from simulation to real. VisDA contains 152,397 synthetic images as the source domain and 55,388 real-world images as the target domain. 12 object categories are shared by these
  two domains. Following previous works \cite{cdan2018, alda2020}, we evaluate all methods on the validation set of VisDA. For VisDA, we made 10000 iterations with a batch size of 72 and $\tau$ was set to 0.25.

  \begin{table}[t]
          \begin{center}
              \begin{tabular}{|@{\hskip3pt}c@{\hskip3pt}|@{\hskip3pt}c@{\hskip3pt}|}
                   \hline
                   Methods & Accuracy (in $\%$)\\
                   \hline
                   \textsc{cdan-e}(*) & 70.1 \\
                   \textsc{alda}(*) & 70.5\\
                   \textsc{djdot}(*) & 68.2 \\
                   \textsc{Robust OT}(*) & 66.3 \\
                   \textsc{djduot} & \textbf{71.9} \\ 
                   \hline
              \end{tabular}
          \end{center}
      \caption{Summary table of DA results on VisDA datasets}
      \label{app:summary_visda}
  \end{table}
}

In this subsection, we detail the setup of our domain adaptation experiments.

{\bfseries Setup.} First note that for all datasets, \textsc{jumbot} uses a stratified sampling on source minibatches as done in \textsc{deepjdot} \cite{deepjdot}. Stratified sampling means that each class has the same number of samples in the minibatches. This is a realistic setting as labels are available in the source dataset.

For Digits datasets, we used the 9 CNN layers architecture and the 1 dense layer classification proposed in \cite{deepjdot}. We trained our neural network on the source domain during 10 epochs before applying \textsc{jumbot}. We used Adam optimier with a learning rate of $2e^{-4}$ with a minibatch size of 500. Regarding competitors, we use the official implementations with the considered architecture and training procedure.

For office-home and VisDA, we employed  ResNet-50 as generator. ResNet-50 is pretrained on ImageNet and our discriminator consists of two fully connected layers with dropout, which is the same as previous works \cite{DANN, cdan2018, alda2020}. As we train the classifier and discriminator from scratch, we set their learning rates to be 10 times that of the generator. We train the model with Stochastic Gradient Descent optimizer with the momentum of $0.9$. We schedule the learning rate with the strategy in \cite{DANN}, it is adjusted by $\chi_p = \frac{\chi_0}{(1+\mu q)^\nu}$, where $q$ is the training progress linearly changing from 0 to 1, $\chi_0 = 0.01$, $\mu = 10$, $\nu = 0.75$. We compare \textsc{jumbot} against recent domain adaptation papers \textsc{dann}\cite{DANN}, \textsc{cdan-e} \cite{cdan2018}, \textsc{alda} \cite{alda2020}, \textsc{deepjdot} \cite{deepjdot} and \textsc{rot} \cite{balaji2020robust} on all considered datasets. We reproduced their scores and on contrary of these papers we do not report the best classification on the test along the iterations but at the end of training, which explains why there might be a difference between reported results and reproduces results. We sincerely believe that the evaluation shall only be done at the end of training as labels are not available in the target domains. But we also report the maximum accuracy along epochs for the Office-Home DA task in table \ref{app:tab_max_oh} and it shows that our method is above all of the competitors by a safe margin of $3\%$.

For Office-Home, we made 10000 iterations with a batch size of 65 and for VisDA, we made 10000 iterations with a batch size of 72. For fair comparison we used our minibatch size and number of iterations to evaluate competitors. The hyperparameters used in our experiments are as follows $\eta_1=0.1, \eta_2=0.1, \eta_3=1, \tau=1, \varepsilon=0.1$ for the digits and for office-home datasets $\eta_1=0.01, \eta_2=0.5, \eta_3=1, \tau=0.5, \varepsilon=0.01$. For VisDA, $\eta_1=0.005, \eta_2=1, \eta_3=1, \varepsilon=0.01$ and $\tau$ was set to 0.3.

\begin{table*}[t]
\centering
\begin{tabular}{llllllllllllll}
 Method & A-C & A-P & A-R & C-A & C-P & C-R & P-A & P-C & P-R & R-A & R-C & R-P & avg\\
 \hline
 \textsc{resnet-50}  & 34.9 & 50.0 & 58.0 & 37.4 & 41.9 & 46.2 & 38.5 & 31.2 & 60.4 & 53.9 & 41.2&  59.9 & 46.1\\
 \textsc{dann}  & 46.2 & 65.2 & 73.0 & 54.0 & 61.0 & 65.2 & 52.0 & 43.6 & 72.0  & 64.7 & 52.3 & 79.2 & 60.7\\
 \textsc{cdan-e} & 52.8 & 71.4 & 76.1 & 59.7 & 70.6 & 71.5 & 59.8 & 50.8 & 77.7 & 71.4& 58.1& \textbf{83.5} & 67.0 \\
\textsc{alda}  & 53.7 & 70.1 & 76.4 & 60.2 & 72.6 & 71.5 & 56.8 & 51.9 & 77.1 & 70.2 & 56.3 & 82.1 & 66.6\\
\textsc{ROT} & 47.2 & 70.8 & 77.6 & 61.3 & 69.9 & 72.0 & 55.4 & 41.4 & 77.6 & 69.9 & 50.4 & 81.5 & 64.6\\
 \textsc{deepjdot}  & 53.4 & 71.7 & 77.2 & 62.8 & 70.2 & 71.4 & 60.2 & 50.2 &77.1 & 67.7 & 56.5 & 80.7 & 66.6\\
 \textsc{jumbot} & \textbf{55.3} & \textbf{75.5} & \textbf{80.8} & \textbf{65.5}  & \textbf{74.4} &  \textbf{74.9}  & \textbf{65.4}  & \textbf{52.7} & \textbf{79.3}& \textbf{74.2} & \textbf{59.9} & 83.4 & \textbf{70.1} \\
\end{tabular}
\caption{Office-Home experiments with maximum classification (ResNet50)}
\label{app:tab_max_oh}
\end{table*}

\subsection{Partial DA}

 For Partial Domain Adaptation, we considered a neural network architecture and a training procedure similar as in the domain adapation experiments which also corresponds to the setting in \cite{liang2020baus}. Our hyperparameters are set as follows : $\tau = 0.06, \eta_1 = 0.003, \eta_2 = 0.75$ and finally $\eta_3$ was set to 10. Regarding training procedure, we made 5000 iterations with a batch size of 65 and for optimization procedure, we used the same as in \cite{liang2020baus}. We do not use the ten crop technic to evaluate our model on the test set as we were not able to reproduce the results from \textsc{ent} and \textsc{pada}. Furthermore, we do not know if the reported results \textsc{ent} and \textsc{pada} were evulated at the end of optimization or during training, but our reported scores are above their scores by at least 5\% on average.

 \subsection{Sensitivity analysis}

\begin{figure*}[t]
    \centering
    \includegraphics[scale=0.45]{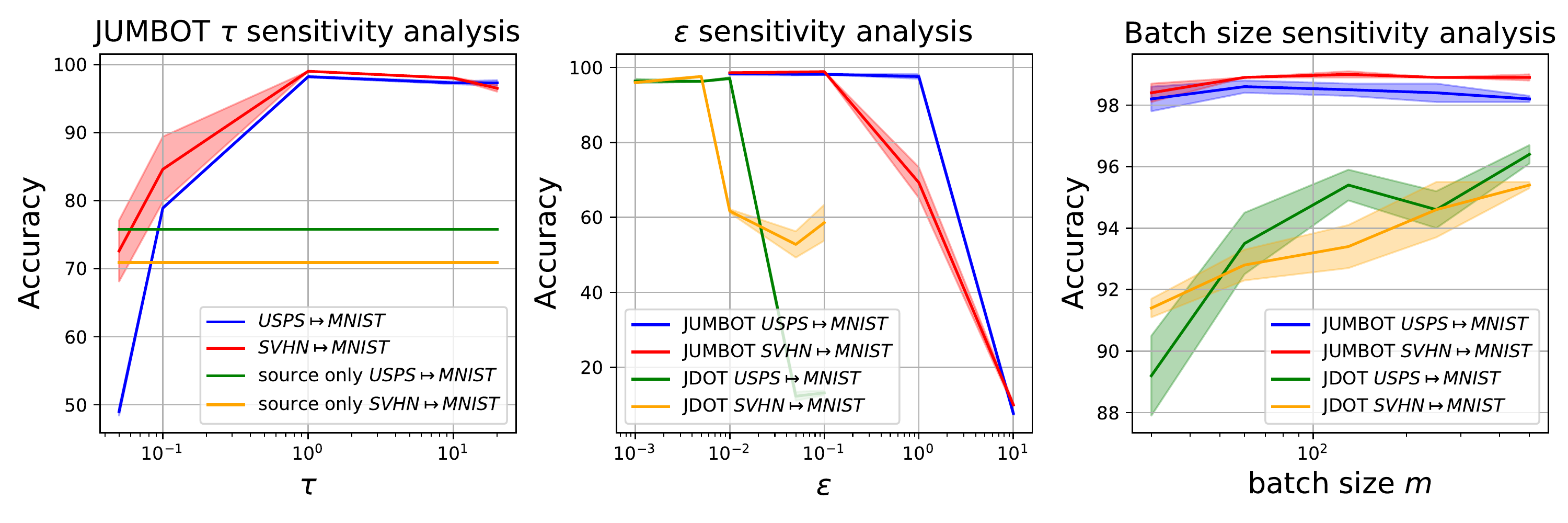}
    \vspace{-0.6cm}
    \caption{(Best viewed in colors) \textsc{deepjdot} and \textsc{jumbot} sensitivity analysis. We report the classification accuracy of \textsc{deepjdot} and \textsc{jumbot} on the DA tasks USPS $\mapsto$ MNIST and SVHN $\mapsto$ MNIST for several hyperparameter variations. We consider the marginal coeffifient $\tau$, the entropic coefficient $\varepsilon$ and the batch size $m$.}
    \label{fig:ablation}
\end{figure*}

 In this paragraph, we make a parameter sensitivity of \textsc{deepjdot} and \textsc{jumbot}. We fix all hyperparameters expect one and we report the accuracy along the variation of the considered hyperparameters. It allows us to see how robust are our method to small perturbations of hyperparameters. We chose to conduct this study on the USPS $\mapsto$ MNIST and SVHN $\mapsto$ MNIST domain adapation tasks. We choose to vary $\tau$ for \textsc{jumbot}, $\varepsilon$ and the batch size for both \textsc{deepjdot} and \textsc{jumbot}. All results are gathered in Figure \ref{fig:ablation}.

 When $\tau$ is too small, \textsc{jumbot} creates negative transfer because of the entropic regularization. When $\tau$ increases, we see that \textsc{jumbot} accuracy increases and it reaches its maximum around $\tau=1$. However when $\tau$ is too high, the marginal distributions are respected and then we see a slight decrease of accuracy due to the OT constraint and the minibatch sampling.

 We now vary $\varepsilon$ for \textsc{jumbot} and the entropic variant of \textsc{deepjdot}. We see that entropy helps getting slightly better results however when the entropic regularization is too high, the accuracy falls. We conjecture that entropic regularized OT regularizes the neural network because the target prediction is matched to a smoothed source label (see a similar discussion in \citep{damodaran_cleot}). And it is well known that label smoothing creates class clusters in the penultimate layer of the neural network \cite{muller_label_smoothin}. 

 Finally we studied the robustness of our method for small batch sizes. While \textsc{jumbot} has a constant accuracy along all batch size, the \textsc{deepjdot} accuracy falls of 4\% for SVHN $\mapsto$ MNIST and 6\% for USPS $\mapsto$ MNIST. The benefits of our method over \textsc{deepjdot} are twofold, it is more robust to small batch sizes and it is performant for small computation budget unlike \textsc{deepjdot}.

 \subsection{Overfitting}

\begin{figure*}[t]
    \centering
    \includegraphics[scale=0.45]{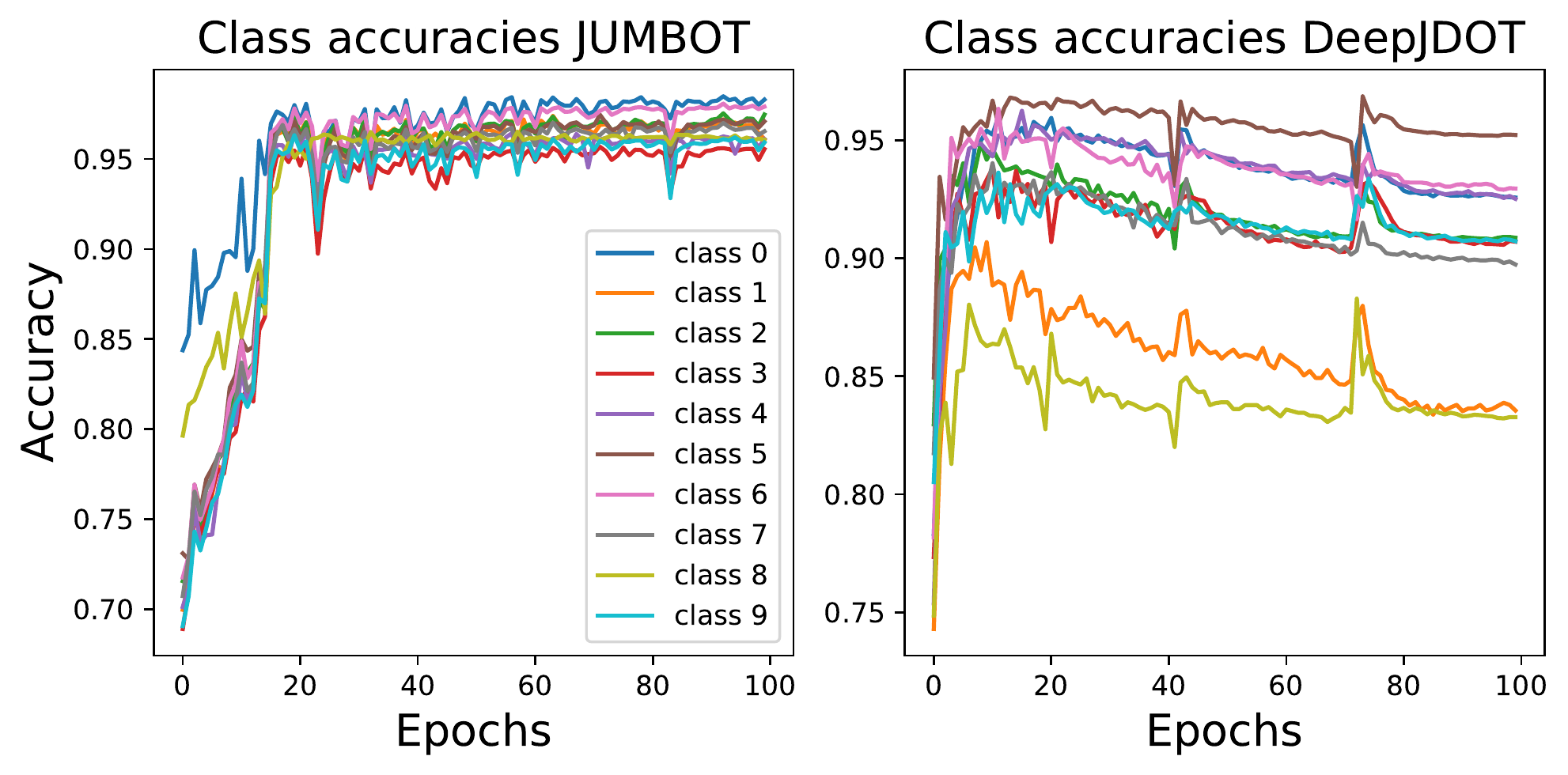}
    \vspace{-0.6cm}
    \caption{(Best viewed in colors) \textsc{deepjdot} and \textsc{jumbot} class accuracies along training. We report the class accuracies along training of \textsc{deepjdot} and \textsc{jumbot} on the DA task MNIST $\mapsto$ M-MNIST for optimal hyper-parameters. Each color represents a different class.}
    \label{fig:overfitting}
\end{figure*}

 In this subsection, we discuss the training behaviour of \textsc{deepjdot} and our method \textsc{jumbot} on the DA task MNIST $\mapsto$ M-MNIST. In Figure \ref{fig:overfitting}, one can see that \textsc{deepjdot} starts overfitting from epoch 30 on each class. There are some classes which are more affected by overfitting than others. The accuracy on each class is reduced of several points. This behaviour is not shared with our method \textsc{jumbot}. Indeed it is more stable, it does not show any sign of overfitting and it has a higher accuracy. This shows the relevance of using our method \textsc{jumbot}.

 \long\def\comment#1{
   \begin{table}[t]
        \begin{center}
            \begin{tabular}{|@{\hskip3pt}c@{\hskip3pt}|@{\hskip3pt}c@{\hskip3pt}|@{\hskip3pt}c@{\hskip3pt}|@{\hskip3pt}c@{\hskip3pt}|@{\hskip3pt}c@{\hskip3pt}|@{\hskip3pt}c@{\hskip3pt}|}
                 \hline
                  $\tau$ & 0.05 & 0.1 & 1 & 10 & 20 \\
                  \hline
                  \textsc{jumbot} & 49.0 $\pm$ 0.6 & 78.9 $\pm$ 0.1 & 98.2 $\pm$ 0.1 & 97.3 $\pm$ 0.2 & 97.3 $\pm$ 0.4 \\
                 \hline
                 $\varepsilon$ & 0.01 & 0.05 & 0.1 & 1 & 10 \\
                 \hline
                  \textsc{jumbot} & 98.4 $\pm$ 0.4 & 98.2 $\pm$ 0.3 & 98.2 $\pm$ 0.1 & 97.6 $\pm$ 0.6 & 7.7 $\pm$ 0.6 \\
                  \hline
                  $\varepsilon$ & 0.001 & 0.005 & 0.01 & 0.05 & 0.1 \\
                 \hline
                  \textsc{jdot} (entropic) & 96.4 $\pm$ 0.6 & 96.3 $\pm$ 0.1 & 97.1 $\pm$ 0.3 & 12.3 $\pm$ 1.2 & 13.2 $\pm$ 0.6 \\
                  \hline
                 batch size & 30 & 60 & 130 & 250 & 500 \\
                 \hline
                  \textsc{jumbot} & 98.2 $\pm$ 0.4 & 98.6 $\pm$ 0.2 & 98.5 $\pm$ 0.2 & 98.4 $\pm$ 0.3 & 98.2 $\pm$ 0.1 \\
                  \textsc{jdot} & 89.2 $\pm$ 1.3 & 93.5 $\pm$ 1.0 & 95.4 $\pm$ 0.5 & 94.6 $\pm$ 0.6 & 96.4 $\pm$ 0.3 \\
                  \hline
            \end{tabular}
        \end{center}
    \caption{Ablation study and parameter sensitivity of \textsc{jumbot} on the USPS $\mapsto$ MNIST domain adapation task. Experiments were run three times.}
    \label{app:ablation}
\end{table}

 \begin{table}[t]
        \begin{center}
            \begin{tabular}{|@{\hskip3pt}c@{\hskip3pt}|@{\hskip3pt}c@{\hskip3pt}|@{\hskip3pt}c@{\hskip3pt}|@{\hskip3pt}c@{\hskip3pt}|@{\hskip3pt}c@{\hskip3pt}|@{\hskip3pt}c@{\hskip3pt}|}
                 \hline
                  $\tau$ & 0.05 & 0.1 & 1 & 10 & 20 \\
                  \hline
                  \textsc{jumbot} & 72.6 $\pm$ 4.5 & 84.6 $\pm$ 4.8 & 99.0 $\pm$ 0.0 & 98.0 $\pm$ 0.1 & 96.5 $\pm$ 0.5 \\
                 \hline
                 $\varepsilon$ & 0.01 & 0.05 & 0.1 & 1 & 10 \\
                 \hline
                  \textsc{jumbot} & 98.6 $\pm$ 0.1 & 98.8 $\pm$ 0.0 & 98.9 $\pm$ 0.1 & 69.3 $\pm$ 4.1 & 10.0 $\pm$ 0.5 \\
                  \hline
                  $\varepsilon$ & 0.001 & 0.005 & 0.01 & 0.05 & 0.1 \\
                 \hline
                  \textsc{jdot} (entropic) & 96.0 $\pm$ 0.2 & 97.6 $\pm$ 0.2 & 61.7 $\pm$ 0.5 & 52.8 $\pm$ 3.5 & 58.6 $\pm$ 4.8 \\
                  \hline
                 batch size & 30 & 60 & 130 & 250 & 500 \\
                 \hline
                  \textsc{jumbot} & ? $\pm$ ? & ? $\pm$ ? & ? $\pm$ ? & ? $\pm$ ? & ? $\pm$ ? \\
                  \textsc{jdot} 91.4 $\pm$ 0.3 & 92.8 $\pm$ 0.5 & 93.4 $\pm$ 0.7 & 94.6 $\pm$ 0.9 & 95.4 $\pm$ 0.1 \\
                  \hline
            \end{tabular}
        \end{center}
    \caption{Ablation study and parameter sensitivity of \textsc{jumbot} on the SVHN $\mapsto$ MNIST domain adapation task. Experiments were run three times.}
    \label{app:ablation}
\end{table}
 }

\end{document}